\PassOptionsToPackage{hyphens}{url}
\documentclass[10pt]{article} 
\usepackage[accepted]{tmlr}

\usepackage{amsmath,amsfonts,bm}

\def\eqref#1{equation~\ref{#1}}

\def\1{\bm{1}}

\def\vs{{\bm{s}}}

\DeclareMathAlphabet{\mathsfit}{\encodingdefault}{\sfdefault}{m}{sl}
\SetMathAlphabet{\mathsfit}{bold}{\encodingdefault}{\sfdefault}{bx}{n}

\usepackage{placeins} 
\usepackage{hyperref}
\usepackage{xspace}
\usepackage{wrapfig}
\usepackage{graphicx}
\usepackage{algorithm}
\usepackage{algorithmic}
\usepackage{enumitem}
\usepackage{comment}
\usepackage{booktabs} 
\usepackage{amsmath} 
\usepackage{amssymb}
\usepackage{multirow} 
\usepackage{arydshln}
\usepackage{amsmath, calc}
\usepackage{array} 
\usepackage{subcaption}

\usepackage{pifont}       
\usepackage{bbding}       
\usepackage{fontawesome}  
\usepackage{colortbl}
\usepackage{tabularx}
\newcolumntype{Y}{>{\centering\arraybackslash}X}
\usepackage{changepage}
\usepackage{anyfontsize}
\newcommand{\shortname}{COLT }

\definecolor{customblue}{RGB}{135,206,250}
\definecolor{mygray}{gray}{0.5}
\usepackage{xspace}
\makeatletter
\DeclareRobustCommand\onedot{\futurelet\@let@token\@onedot}
\def\@onedot{\ifx\@let@token.\else.\null\fi\xspace}
\def\eg{\emph{e.g}\onedot} 
\def\ie{\emph{i.e}\onedot} 
\def\cf{\emph{c.f}\onedot} 
 \def\vs{\emph{vs}\onedot}

\makeatother

\definecolor{mygreen}{RGB}{0,153,86}

\newcolumntype{x}[1]{>{\centering\arraybackslash}p{#1pt}}

\newlength\savewidth

\usepackage{newfloat}
\usepackage{listings}
\DeclareCaptionStyle{ruled}{labelfont=normalfont,labelsep=colon,strut=off} 
\floatstyle{ruled}
\newfloat{listing}{tb}{lst}{}
\floatname{listing}{Listing}
\title{COLT: Enhancing Video Large Language Models \\ with Continual Tool Usage}

\author{
\name\hspace{-0.4em}Yuyang Liu\thanks{Equal contribution.} \email yuyang.liu@mbzuai.ac.ae  \\
\addr Department of Computer Vision\\
University of Mohamed bin Zayed University of Artificial Intelligence
\AND
\name Meng Cao\footnotemark[1] \email mengcaopku@gmail.com \\
\addr Department of Computer Vision\\
University of Mohamed bin Zayed University of Artificial Intelligence
\AND
\name Xinyuan Shi \email xinyuan.shi@mbzuai.ac.ae \\
\addr Department of Computer Vision\\
University of Mohamed bin Zayed University of Artificial Intelligence
\AND
\name Xiaodan Liang \email xiaodan.liang@mbzuai.ac.ae\\
\addr Department of Computer Vision\\
University of Mohamed bin Zayed University of Artificial Intelligence
}

\def\month{01}  
\def\year{2026} 
\def\openreview{\url{https://openreview.net/forum?id=NT9tHHTlXn}} 

\begin{document}

\maketitle

\begin{abstract}
The success of Large Language Models (LLMs) has significantly propelled the research of video understanding. To leverage the strengths of specialist models (\ie, tool) for specific video tasks, recent video LLMs have focused on integrating specialist models as tool usage capabilities into their architectures. Existing methods either prompt closed-source LLMs or employ the instruction tuning paradigm for tool-use finetuning. These methods, however, assume an established repository of \emph{fixed} tools and struggle to generalize to real-world environments where tool data is perpetually evolving and streaming in. To this end, we propose to enhance open-source video LLMs with COntinuaL Tool usage (termed COLT), which automatically acquires tool-use ability in a successive tool stream without suffering ``catastrophic forgetting'' of the past learned tools. Specifically, our COLT incorporates a learnable tool codebook as a tool-specific memory system. Then, relevant tools are dynamically selected based on the similarity between user instructions and tool features within the codebook. To unleash the tool usage potential of video LLMs, we collect a video-centric tool-use instruction tuning dataset VideoTool. Extensive experiments on both previous video LLM benchmarks and the tool-use-specific VideoTool test split demonstrate the state-of-the-art performance of our proposed COLT.
\end{abstract}

\section{Introduction}

\begin{figure}[t] 
         \hspace{-18mm}
         \resizebox{1.25\textwidth}{!}{
        \includegraphics[trim={10 20 10 10}, clip]{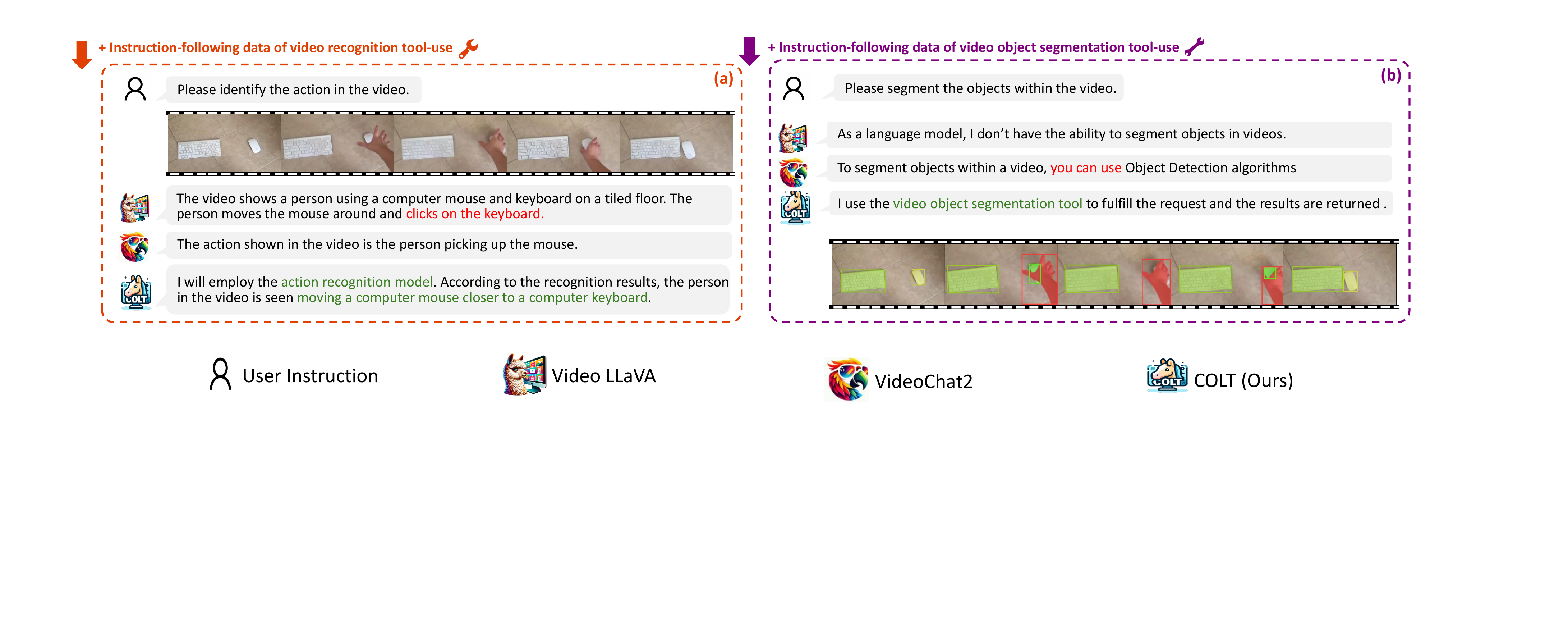}
    }
         \vspace{-32mm}
        \caption{Our proposed COLT continually learns to invoke tools from tool-stream data without catastrophic forgetting. Benefiting from tool usage, COLT (a) is adept at dynamic content understanding and (b) supports flexible generation compared to existing methods \cite{lin2023video,li2023mvbench}. The incorrect parts of responses are marked in red.}
	\label{fig:teaser}
	\vspace{-3mm}
\end{figure}

Large Language Models (LLMs), \eg, GPT~\citep{achiam2023gpt,brown2020language}, PaLM~\citep{anil2023palm}, and LLaMA~\citep{touvron2023llama,touvron2023llama2}, have exhibited remarkable success in understanding user instructions, aligning with human intents, and generating trustworthy responses.
Trained on large-scale corpora and equipped with billions of parameters, these models demonstrate impressive generalization abilities across a wide range of natural language tasks such as reasoning, summarization, and question answering.
They have fundamentally reshaped the paradigm of natural language understanding by enabling unified instruction-following interfaces that support various downstream applications, from conversational agents to coding assistants.
Recently, the research focus has been gradually shifting from pure text-based LLMs to multi-modal LLMs such as BLIP-2~\citep{li2023blip}, Flamingo~\citep{alayrac2022flamingo}, and MiniGPT-4~\citep{zhu2023minigpt}, which augment language models with the capability to perceive and reason over multiple modalities such as vision, audio, and video.
By integrating visual encoders with pretrained LLMs, these models can interpret images or videos and produce context-aware, grounded responses, thus bridging perception and language understanding.
Such a shift extends the role of LLMs from passive text generators to more general perceptual agents, which has aroused significant research interest within the academic community and beyond.
Despite the progress in image-based LLMs, advancing LLMs' capacity to comprehend video data remains a demanding pursuit. 
Recent representative efforts such as Video-LLaMA~\citep{zhang2023video} and VideoChat~\citep{li2023videochat} illustrate the potential of video LLMs, yet the field is still in its early stage. 
The overwhelming majority of video LLM methods follow the instruction-tuning paradigm. 
Building upon open-source LLMs such as Vicuna~\citep{vicuna2023} and LLaMA~\citep{touvron2023llama,touvron2023llama2}, they entail end-to-end training on instruction-tuning datasets generated by GPT models~\citep{achiam2023gpt,brown2020language}.
These methods, however, demonstrate limited ability to comprehend video actions and present responses with hallucinations (\eg, non-existent ``keyboard click'' actions). Besides, most of them lack the capability of \emph{any-to-any} generation, \eg, video object segmentation in Figure \ref{fig:teaser} (b).

\begin{wrapfigure}[18]{r}{0.48\textwidth} 
    \vspace{-4mm} 
    \centering
    \includegraphics[width=\linewidth]{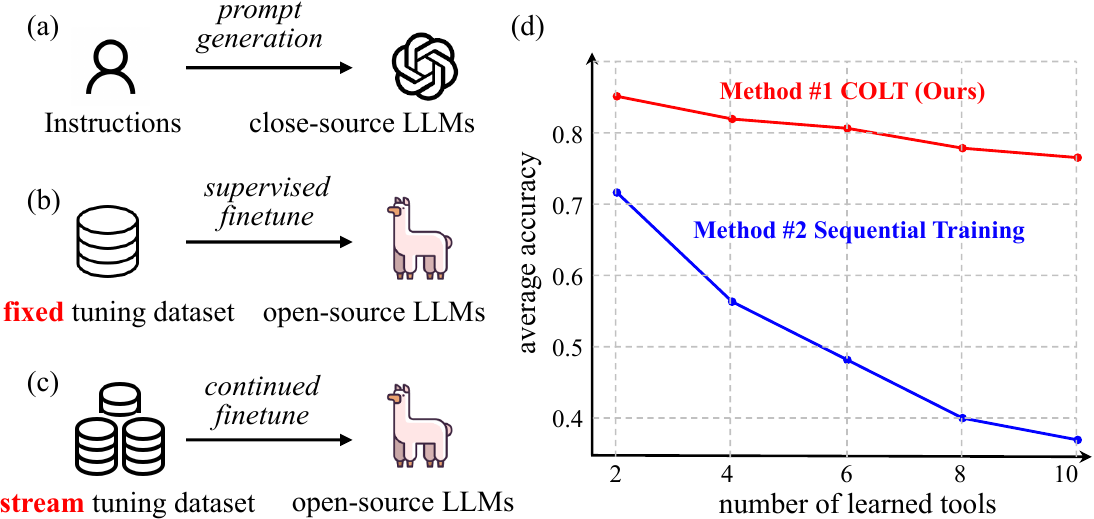}
    \vspace{-4mm}
    \caption{
        (a) \textbf{Agent-based methods} bootstrap closed-source LLMs via delicately designed prompts; 
        (b) Instruction tuning with \textbf{fixed} tool-use dataset; 
        (c) Instruction tuning with \textbf{stream} tool-use dataset (Ours); 
        (d) \textbf{Average tool calling accuracy on VideoTool \vs\  learned tools.}
        Sequential training denotes training on a sequence of tasks independently.
    }
    \label{fig:teaserMotivate}
    \vspace{0mm}
\end{wrapfigure}

To alleviate this, agent-based methods such as AssistGPT~\citep{gao2023assistgpt}, ToolLLM~\citep{qin2023toolllm}, DoraemonGPT~\citep{yang2024doraemongpt}, VideoAgent~\citep{fan2024videoagent}, and VideoAgent-2~\citep{wang2024videoagent} 
are introduced to advocate the tool-use capabilities of LLMs. More recently, VITAL~\citep{zhang2025thinkingvideosmultimodaltoolaugmented} explores reinforcing tool-augmented video reasoning via multimodal reinforcement learning on long videos.
As shown in Figure~\ref{fig:teaserMotivate}~(a), 
they mostly bootstrap closed-source LLMs~\citep{achiam2023gpt,brown2020language} 
to decompose the user instructions into more manageable sub-tasks and incorporate external tools to assist. 
These training-free methods require delicately designed prompts and lengthy context windows for instructions~\citep{liu2023llava}. 
Instead, recent attempts like LLaVA-plus~\citep{liu2023llava} and ToolLLM~\citep{qin2023toolllm} 
explore developing tool-specific instruction tuning data and elicit the tool-use capabilities of open-source LLMs 
(\cf Figure~\ref{fig:teaserMotivate}~(b)). 
However, these methods typically assume a pre-set repository containing fixed tools, 
which is difficult to generalize into the real-world scenario where tool data arrives in a never-ending stream. 
In such cases, the conventional sequential training, \ie, training on new data with the pre-trained weights on the previous tool data as initialization, 
leads to significant catastrophic forgetting of prior tool-usage knowledge. 
For example in Figure~\ref{fig:teaserMotivate}~(d), 
the tool-call accuracy of sequential training rapidly decreases as the number of tools learned increases.
In this paper, we propose to enhance open-source video LLMs with \textbf{C}\textbf{O}ntinua\textbf{L} \textbf{T}ool usage (dubbed as \textbf{COLT}). By ``continual'', we mean that video LLMs possess the inherent capacity to learn and automatically activate tool functionality in a successive tool-stream format. To achieve this, two critical issues arise: 1) \emph{how to build} a tool-specific memory to mitigate catastrophic forgetting of past knowledge? One intuitive idea is to store a few past tool data and replay them with the new tool data for continual training. However, the requisite memory expands proportionally with the increase in the number of tools, and previous data may be unavailable due to privacy constraints. Instead, we set up a \emph{tool codebook} consisting of learnable prompts to store tool-specific information in a more concise manner; 2) \emph{how to exploit} relevant tools based on input user instructions? For tool activation, we compute the cosine similarity between the embedded user instructions and each prompt within the tool codebook. The highest-response prompts are then selected to ``instruct'' the model to invoke appropriate tools.

Although several video LLM benchmarks \citep{maaz2023video,li2023mvbench} have been proposed, the community still lacks a benchmark for video-centric tool-use instruction tuning. To this end, we collect \textbf{VideoTool} to facilitate tool-use capabilities for open-source video LLMs. Specifically, we collect a set of video specialist models and prompt GPT \citep{achiam2023gpt} to generate diverse instructions for tool calls. Then the tool execution results are used to form the instruction-following dataset. Under our continual learning setup, COLT can incrementally incorporate newly added tools from a expanding repository, while mitigating forgetting of previously learned ones.

In summary, our contributions are threefold:
\begin{itemize}[topsep=0pt, partopsep=0pt, leftmargin=13pt, parsep=0pt, itemsep=3pt]
    \item We proposed COLT, a video LLM with continual tool usage. By maintaining a tool codebook, COLT incrementally learns new tools without catastrophic forgetting.
    \item A tool-using instruction-following dataset VideoTool is introduced to unlock the potential for tool usage within video LLMs.    
    \item Experiments on both existing video LLM benchmarks and the VideoTool test split have manifested the state-of-the-art performance of COLT. For example, on MSRVTT-QA \citep{chen2011collecting}, our COLT boosts the previous SOTA method \citep{li2023mvbench} by 8.2\% on the accuracy of zero-shot video-question answering.
\end{itemize}

\section{Related Work}
\label{rela_work}

\textbf{Video LLMs.} Recent successes of LLMs \citep{achiam2023gpt,anil2023palm} have shed light on the burgeoning proliferation of video LLMs \citep{lin2023video,maaz2023video,gao2023assistgpt}. The mainstream video LLMs follow the instruction-tuning paradigm. Building upon open-source LLMs \citep{vicuna2023,touvron2023llama,touvron2023llama2}, this kind of method adapts the pre-trained video features into LLM understandable representations via multi-layer perception projector \citep{lin2023video,maaz2023video,li2023llama,munasinghe2023pg,liu2024st,jin2023chat,liu2023one}, Q-former \citep{zhang2023video,li2023videochat}, or discretization tokenizer \citep{jin2024video}. These methods, however, fail to generalize to broader video understanding tasks, which may require flexible input and output formats \citep{wu2023next,zhan2024anygpt}. Several methods attempt to achieve this by employing additional functional modules \citep{jin2024video,munasinghe2023pg} (\eg, grounding \citep{munasinghe2023pg} or diffusion modules \citep{jin2024video}), which are not flexible enough to adapt to diverse video understanding needs. To this end, another stream of agent-based methods \citep{qin2023toolllm,yang2024doraemongpt,fan2024videoagent,wang2024videoagent} is proposed, where multi-modal agents mostly bootstrap closed-source LLM \citep{achiam2023gpt,brown2020language} to decompose solution paths and call external tools. This kind of method relies heavily on delicately designed prompts and may fail to acquire the usage of a novel tool. Therefore, how to implement automatic invocation of related tools in video LLMs under a non-stationary tool stream remains to be solved. This paper combines the strengths of the above two methodologies to empower video LLMs with automatic and continual tool-use ability. 

\noindent \textbf{Continual Learning.} Continual learning \citep{wang2024comprehensive,lee2017overcoming,mccloskey1989catastrophic} refers to the ability to incrementally acquire, update, and accumulate knowledge throughout the model lifetime without catastrophically forgetting previously learned information. 
Conceptually, four varieties of methodologies are posited. Regularization-based approaches \citep{kirkpatrick2017overcoming,li2017learning,feng2022overcoming,yang2024CLL-CLIP} strike the balance between the old and new tasks by adding explicit regularization terms.
Architecture-based approaches ~\citep{yoon2018DEN,li2019Learn2Grow,ke2020CAT} isolate model parameters for different tasks.
Rehearsal-based methods \citep{bonicelli2022effectiveness,chen2023dynamic,lin2023pcr} typically use a memory buffer to store several training samples from previous classes, which are used to approximate and recover old data distributions. Prompt-based methods \citep{wang2022learning,smith2023coda,wang2022dualprompt,li2024kc} usually construct task-adaptive prompts and select appropriate prompts during inference. This kind of method is rehearsal-free and thus more computationally efficient. L2P \citep{wang2022learning} introduces the concept of a prompt pool and selects prompts by a query-key mechanism. To overcome the separate optimization issue of L2P \citep{wang2022learning}, CODA-Prompt \citep{smith2023coda} assembles learnable prompts with input-conditioned weights. DualPrompt \citep{wang2022dualprompt} and KC-Prompt \citep{li2024kc} set up prompt pools to respectively encode task-invariant and task-specific knowledge. 

Borrowing the favorable rehearsal-free merit, our COLT inventively devises a video LLM with the continual tool-use learning capability. We differ from current prompt-based methods \citep{wang2022learning,smith2023coda,wang2022dualprompt,li2024kc} in the following two aspects. Firstly, our focus is on the more complex task of multi-modal language generation as opposed to the basic image classification task. Secondly, our COLT employs straight-through \citep{van2017neural,bengio2013estimating} gradient estimation to mitigate the optimization challenges encountered in previous prompt-based approaches \citep{wang2022learning,smith2023coda,wang2022dualprompt,li2024kc}.

\section{Dataset Construction}
\label{dataset}

\textbf{Dataset Structure.} Each instance in VideoTool contains two rounds of conversations between \texttt{human} and \texttt{gpt}. The first round includes human instructions related to the video content and the LLM responses on choosing appropriate tools; The second round contains the execution results of the corresponding tools and the final responses from LLMs. Specifically, we follow \citep{yao2022react,yang2023mm,liu2023llava} to unify the response format of GPT into three fields, including \texttt{thought}, \texttt{action} , and \texttt{value} to mimic a human-like task-solving procedure. 

\begin{table}[t]
    \small
    \caption{\textbf{Tool repository} of the proposed VideoTool including single and compositional tools.}
    \vspace{-2mm}
    \centering
    \renewcommand{\arraystretch}{1.05}
    \begin{tabularx}{\linewidth}{c|Y|Y}
        \toprule
        & \textbf{Tool} & \textbf{Specialist Model} \\
        \midrule
        \multirow{8}{*}{\rotatebox{90}{\footnotesize \it single tool}}
        & Action Recognition (AR) & VideoMAE \cite{tong2022videomae} \\
        & Dense Video Caption (DVC) & PDVC \cite{wang2021end} \\
        & Temporal Action Localization (TAL) & InternVideo \cite{wang2022internvideo} \\
        & Optical Character Recognition (OCR) & EasyOCR~\citep{EasyOCR} \\
        & Automatic Speech Recognition (ASR) & Whisper \cite{radford2023robust} \\
        & Video Relation Detection (VRD) & VidVRD \cite{shang2017video} \\
        & Video Object Segmentation (VOS) & VisTR~\cite{wang2021end} \\
        & Text-to-Video Generation (T2V) & T2V \cite{khachatryan2023text2video} \\
        \midrule
        \multirow{2}{*}{\rotatebox{90}{\footnotesize \it multi.}}
        & AR + ASR & VideoMAE + Whisper \\
        & AR + VOS & VideoMAE + VisTR \\
        \bottomrule
    \end{tabularx}
    \label{table:toollist}
\end{table}

\textbf{Dataset Construction.} This dataset is constructed using GPT-3.5-turbo \citep{achiam2023gpt} with self-instruction. The involved tools\footnote{In this paper, we use \emph{tool} to represent the general skills and \emph{specialist model} to denote the specific model. \label{tooldef}} incorporate both video understanding and generation tasks. The prompt and in-context learning cases are available in supplementary materials. As shown in Table \ref{table:toollist}, we collect ten tools including eight single tool and two compositional tool. The tool list can be easily extended using a similar dataset construction manner. We initially generate 5,000 instruction-following samples for each tool. We conduct both data format checks and manual verification of semantic meanings to filter out error data. After a thorough examination, the dataset is partitioned into distinct training and testing splits, adhering to a proportionality ratio of 9:1. The full statistics of the final VideoTool are summarized in supplementary materials. 

\section{Method}
\label{Method}

Our COLT learns the continual tool usage from the stream instruction-tuning dataset $\{\mathcal{D}^t\}_{t=1}^T$, each $\mathcal{D}^t$ containing triplets of the visual data $\mathbf{X}_{\mathbf{v}}^t$, user instructions $\mathbf{X}_{\mathbf{w}}^t$, and the response sequence $\mathbf{X}_{\mathbf{a}}^t$. For clarity, we elaborate on the architecture and training strategy for one single dataset $\mathcal{D}^t$ and the superscripts of $\mathbf{X}_{\mathbf{v}}^t$, $\mathbf{X}_{\mathbf{w}}^t$, and $\mathbf{X}_{\mathbf{a}}^t$ are omitted.


\begin{figure*}[t]
	\centering
         \includegraphics[width=0.95\textwidth]{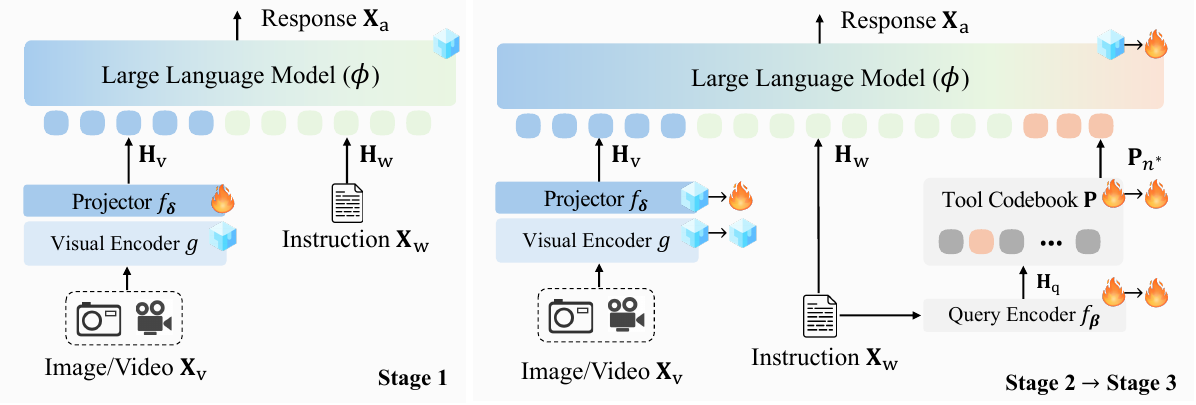}
         \vspace{-1mm}
	\caption{\textbf{An overview of COLT.} Stage 1 aligns the visual and textual modalities through the individual training of the linear projector ${f}_{\boldsymbol{\delta}}$; In stage 2 and stage 3, the prompt within tool codebook $\mathbf{P}$ is adaptively selected according to the cosine similarity with the query feature $\mathbf{H}_{\mathbf{q}}$.}
	\label{fig:pipeline}
	\vspace{-2mm}
\end{figure*}

\subsection{Architecture}  \label{sec:4_1} 

\textbf{Visual \& Textual Embedding.} Following \citep{lin2023video}, we adopt LanguageBind  vision encoder \citep{zhu2023languagebind} ${g}(\cdot)$  to extract visual features. Then, a linear projection layer ${f}_{\boldsymbol{\delta}}(\cdot)$ parameterized by $\boldsymbol{\delta}$ is applied to compress the visual information into an LLM understandable feature space. 
\begin{equation}
\mathbf{H}_{\mathbf{v}} = {f}_{\boldsymbol{\delta}}  ({g} (\mathbf{X}_{\mathbf{v}})), 
\end{equation}
where $\mathbf{H}_{\mathbf{v}} \in \mathbb{R}^{P \times C}$ denotes the visual embeddings, with $P$ and $C$ respectively representing the patch number and the feature dimension. For input user instructions $\mathbf{X}_{\mathbf{w}}$, we adopt the widely used BPE tokenizer \citep{sennrich2016neural} to obtain the textual embeddings $\mathbf{H}_{\mathbf{w}} \in \mathbb{R}^{S \times C}$, where $S$ is the textual token number.

\textbf{Tool Codebook \& Query Encoder.} As shown in Figure \ref{fig:pipeline}, we set up a tool codebook $\mathbf{P}$ as the tool-specific memory for continual tool usage. The codebook consists of $N$ learnable tool prompts, \ie, $\mathbf{P} = \{\mathbf{P}_{n}\}_{n=1}^{N}$ with $\mathbf{P}_{n} \in \mathbb{R}^{C \times 1}$. For each input user instruction $\mathbf{X}_{\mathbf{q}}$, we aim to retrieve relevant tool prompts from the codebook. To achieve this, a \emph{query} encoder ${f}_{\boldsymbol{\beta}}$ with parameters $\boldsymbol{\beta}$ is firstly employed to encode user instructions $\mathbf{X}_{\mathbf{q}}$ into $\mathbf{H}_{\mathbf{q}} \in \mathbb{R}^{C \times 1}$. Then, we compute the cosine similarities between the query embedding $\mathbf{H}_{\mathbf{q}}$ and each tool prompt $\mathbf{P}_{n}$, $n \in \{1, 2, \cdots, N\}$. The tool prompts with the top-$K$ highest similarity scores are selected.
\begin{equation}
n^*={\arg \operatorname{topk}}_{n \in[1, N]} \; (\mathbf{P}_{n}^\top \cdot \mathbf{H}_{\mathbf{q}}),
\label{eq:promptSelect}
\end{equation}
where $n^{*}$ is the selected tool prompt index. Then each $\mathbf{P}_{n^{*}}$ is concatenated with the video feature $\mathbf{H}_{\mathbf{v}}$ and user instruction features $\mathbf{H}_{\mathbf{w}}$ and fed into LLMs. The hint contained in the codebook helps alleviate the catastrophic forgetting when facing the tool stream data. 

\subsection{Training} \label{sec:4_2} 

The proposed COLT adopts a three-stage training methodology, with each stage featuring unique trainable parameters $\boldsymbol{\theta}$ and loss functions.

\textbf{Stage 1: Video-to-text Alignment.} This stage aims to train the projection layer ${f}_{\boldsymbol{\delta}}$ parameterized by $\boldsymbol{\delta}$, which acts as a visual tokenizer to align visual signals with pre-trained LLM word embedding. As shown in Figure \ref{fig:pipeline}, we freeze all the weights except the projection layer in this stage, \ie, trainable parameters $\boldsymbol{\theta} = \boldsymbol{\delta}$. 
Given the predicted response sequence $\mathbf{X}_{\mathbf{a}} = \{\mathbf{X}_{\mathbf{a}}^{1}, \mathbf{X}_{\mathbf{a}}^{2}, \cdots \mathbf{X}_{\mathbf{a}}^{L}\}$ of length $L$, we use the vanilla auto-regressive language modeling (LM) loss to supervise this stage of training:
\begin{equation}
\setlength{\abovedisplayskip}{3pt}
\setlength{\belowdisplayskip}{3pt} 
\mathcal{L} =\mathcal{L}_{\text {LM}} =-\frac{1}{L} \sum_{i=1}^L \; \log p_{\boldsymbol{\theta}} (\mathbf{X}_{\mathbf{a}}^{i} \mid \mathbf{H}_{\mathbf{v}}, \mathbf{H}_{\mathbf{w}}, \mathbf{X}_{\mathbf{a}}^{<i}),
\label{eq:lossstage1}
\end{equation}
where $\mathbf{X}_{\mathbf{a}}^{<i}$ is the response tokens before the $i$-th token.

\textbf{Stage 2: Tool Codebook Pre-training.} In this stage, we pre-train the tool codebook $\mathbf{P}$ and the query encoder ${f}_{\boldsymbol{\beta}}$ while keeping the other parts frozen, \ie, the trainable parameters are $\boldsymbol{\theta} = \{\mathbf{P}, \boldsymbol{\beta}\}$. We firstly select the appropriate tool prompt  $\mathbf{P}_{n^{*}}$ via Eq.\eqref{eq:promptSelect}. 
Then $\mathbf{P}_{n^{*}}$ serves as an additional condition for language modeling as follows: 
\begin{equation}
\small
\setlength{\abovedisplayskip}{3pt}
\setlength{\belowdisplayskip}{3pt}
\mathcal{L}_{\text {LM}}' =-\frac{1}{L} \sum_{i=1}^L \; \log  p_{\boldsymbol{\theta}} (\mathbf{X}_{\mathbf{a}}^{i} \mid \mathbf{H}_{\mathbf{v}}, \mathbf{H}_{\mathbf{w}}, {\color{mygreen}\mathbf{P}_{n^{*}}}, \mathbf{X}_{\mathbf{a}}^{<i}),
\label{eq:lm_new}
\end{equation}
where $\mathbf{X}_{\mathbf{a}}^{i}$ and $\mathbf{X}_{\mathbf{a}}^{<i}$ is as defined in Eq.~\eqref{eq:lossstage1}.
However, directly optimizing $\mathcal{L}_{\text {LM}}'$ will truncate the gradients w.r.t the query encoder ${f}_{\boldsymbol{\beta}}$ due to the non-derivable $\arg \operatorname{topk}$ operation in Eq. \eqref{eq:promptSelect}.
Therefore, we resort to the straight-through estimator \citep{van2017neural,bengio2013estimating} to approximate the gradient computation and define the overall loss function as follows:
\begin{equation}
\small
\setlength{\abovedisplayskip}{3pt}
\setlength{\belowdisplayskip}{3pt}
\mathcal{L} 
= \mathcal{L}_{\text {LM}}' 
+ \lambda_{1} \underbrace{\left\|\operatorname{sg}\left[\mathbf{H}_{\mathbf{w}}\right]-\mathbf{P}_{n^{*}}\right\|_2^2}_{\mathcal{L}_{q}}+\lambda_{2} \underbrace{\left\|\mathbf{H}_{\mathbf{w}}-\operatorname{sg}[\mathbf{P}_{n^{*}}]\right\|_2^2}_{\mathcal{L}_{c}},
\label{eq:lossstage2}
\end{equation}
where $\operatorname{sg}[\cdot]$ stands for the stop-gradient operator that acts as an identity in the forward process and has zero partial derivatives during backward propagation.
$\mathcal{L}_{q}$ is the quantisation loss for codebook update by forcing $\mathbf{P}$ towards the user instruction embeddings $\mathbf{H}_{\mathbf{w}}$, $\mathcal{L}_{c}$ is the commitment loss to prevent unrestricted update of codebook embeddings, $\lambda_{1}$ and $\lambda_{2}$ are the balancing weights.

\textbf{Stage 3: End-to-end Fine-tuning.} We keep the vision encoder ${g}(\cdot)$ frozen and finetune remaining parts including the projection layer ${f}_{\boldsymbol{\delta}}$, tool codebook $\mathbf{P}$, textual encoder ${f}_{\boldsymbol{\beta}}$ and LLM parameterized by $\boldsymbol{\phi}$, \ie, the trainable parameters are $\boldsymbol{\theta} = \{\boldsymbol{\delta}, \mathbf{P}, \boldsymbol{\beta}, \boldsymbol{\phi}\}$. The training loss is the same as Eq. \eqref{eq:lossstage2}.

\section{Experiments}

\subsection{Experimental Settings}\label{sec:5_1}

\noindent \textbf{Training Datasets.} 
In the first stage, we follow Video LLaVA \cite{lin2023video} to use LAION-CC-SBU image subset \cite{schuhmann2021laion} and the filtered CC3M video dataset \cite{changpinyo2021conceptual} for video-to-text alignment. For the second and third stage, we use the combined dataset of 665K image-text instruction data from LLaVA v1.5 \cite{liu2023improved} and 100K video-text instruction data from Video-ChatGPT \cite{maaz2023video}. The tool-use instruction data of our collected VideoTool is introduced into continual training sequentially. To ensure conformity in data formatting, we reformat all the instruction tuning datasets to the \texttt{thought}-\texttt{action}-\texttt{value} pattern. Refer to supplementary materials for more details.

\begin{table*}[t]
	\small
	\setlength{\abovecaptionskip}{-0.05cm}
	\setlength{\belowcaptionskip}{-0.2cm}
	\caption{\textbf{Comparisons with state-of-the-art methods on zero-shot video-question answering.} ``Acc." denotes accuracy (\%) and ``Score" denotes the relative score from 0 to 5 assigned by GPT \cite{brown2020language}. The best performance is in \textbf{bold} and the second best is \underline{underlined}. The backend LLMs include Vicuna-7B \cite{vicuna2023} and LLaMA-7B \cite{touvron2023llama}.} 
	\label{tab:zeroqa}
	\renewcommand\arraystretch{1.1}
	\begin{center}
			\begin{tabular}{lrccccccc}
				\toprule
				\multirow{2}{*}{\textbf{Method}} &  \multirow{2}{*}{\textbf{LLM}} & \multicolumn{2}{c}{\textbf{MSVD-QA}} & \multicolumn{2}{c}{\textbf{MSRVTT-QA}} & \multicolumn{2}{c}{\textbf{ActivityNet-QA}}  \\
				\cmidrule(lr){3-4} \cmidrule(lr){5-6} \cmidrule(l){7-8}&  & \textbf{Acc.} & \textbf{Score} & \textbf{Acc.} & \textbf{Score} & \textbf{Acc.} & \textbf{Score}  \\
				\midrule
				Video-LLaMA  \cite{zhang2023video} & Vicuna-7B  & 51.6 & 2.5 & 29.6 & 1.8 & 12.4 & 1.1  \\
				VideoChat   \cite{li2023videochat} & Vicuna-7B   & 56.3 & 2.8 & 45.0 & 2.5 & 26.5 & 2.2   \\
				Video-ChatGPT  \cite{maaz2023video} & Vicuna-7B  & 64.9 & 3.3 & 49.3 & 2.8 & 35.2 & 2.7  \\
				BT-Adapter   \cite{liu2023one} & Vicuna-7B  & 67.5 & 3.7  & 57.0 & 3.2 & 45.7 & 3.2  \\
				LLaMA-VID  \cite{li2023llama} & Vicuna-7B  & 69.7 & 3.7 & 57.7 & 3.2 & 47.4 & 3.3  \\
				LLaMA-VID   \cite{li2023llama} & Vicuna-13B  & 70.0 & 3.7 & 58.9 & 3.3 & 47.5 & 3.3 \\
				Video-LLaVA   \cite{lin2023video} & Vicuna-7B  & 70.7 & {3.9} & 59.2 & 3.5 & 45.3 & 3.3 \\
				Chat-UniVi \cite{jin2023chat} & Vicuna-7B  & 65.0 & 3.6 & 54.6 & 3.1 & 45.8 & 3.2  \\
				LLaMA-Adapter   \cite{zhang2023llama} & LLaMA-7B  & 54.9 & 3.1 & 43.8 & 2.7 & 34.2 & 2.7   \\
				VideoChat2   \cite{li2023mvbench} & Vicuna-7B   & 70.0  & {3.9} & 54.1 & 3.3 & 49.1 & 3.3  \\
				ST-LLM   \cite{liu2024st}  & Vicuna-7B  & 74.6  & 3.9 & 63.2 & 3.4 & 50.9 & 3.3  \\
    
				\rowcolor{customblue!20}  \shortname$_\text{joint}$ (Ours)  & Vicuna-7B  & \textbf{78.2} & \textbf{4.2} &  \textbf{65.1} & \textbf{3.6} & \textbf{54.7} & \textbf{3.8} \\
				\rowcolor{customblue!20} \shortname$_\text{5$\times$2}$ (Ours)  & Vicuna-7B  & \underline{75.5}  & \underline{3.9} & \underline{63.9} & \underline{3.5} & \underline{52.5} & \underline{3.5}   \\
				\rowcolor{customblue!20} \shortname$_\text{10$\times$1}$ (Ours)  & Vicuna-7B  & 74.7  & 3.9 & 63.0 & 3.4 & 51.2 & 3.4   \\
				\bottomrule
			\end{tabular}
	\end{center}
\end{table*}

\noindent \textbf{Implementation Details.} For vision encoder $\boldsymbol{g}(\cdot)$, we choose pre-trained LanguageBind \cite{zhu2023languagebind} with ViT-L/14 \cite{dosovitskiy2020image}. The text tokenizer is derived from LLaMA \cite{touvron2023llama} with a vocabulary size of 32,000, and Vicuna-7B v1.5 \cite{vicuna2023} is employed as the large language model. We uniformly sample 8 frames from each video, and each frame is resized to 224 $\times$ 224. We set the batch size to 256 for the first stage and 128 for the second and third stages. AdamW optimizer is used with a cosine decay schedule. We set learning rate to $1 \times 10^{-4}$, $1 \times 10^{-4}$, and $1 \times 10^{-5}$ for three stage training, respectively. The balancing weight $\lambda_1$ and $\lambda_2$ in Eq. \eqref{eq:lossstage2} are set to 1 and 0.25. The codebook size $N$ and selected prompt number $K$ are set to 50 and 3, respectively. COLT is trained on 8 NVIDIA A100 GPUs (80 GB memory each), and the full training process takes approximately 10 hours.

\textbf{Evaluation Metrics of Tool Continual Learning.} We detail the metrics of average accuracy and average forgetting used in Sec.~\ref{sec:5_3}.

Let $\alpha_{k,j} \in [0, 1]$ denote the tool call accuracy of $j$-th tool after incrementally training on the $k$ tool data ($j \leq k$). The \emph{average accuracy} of a specific tool denotes the overall call accuracy during the incremental learning process. Then the average accuracy at task $k$ is defined as follows.
\begin{equation}
\text{AA}_{k}=\frac{1}{k} \sum_{j=1}^k a_{k, j}.
\end{equation}
Since average accuracy does not provide any information about the forgetting profile of the continual learning process, \emph{average forgetting} is introduced to bridge the gap. For a particular tool, the forgetting measure is defined as the difference between the maximum tool call accuracy throughout the past learning process and the current tool call accuracy. In particular, the forgetting for the $j$-th tool after incrementally training up to $k$ tool is as follows.
\begin{equation}
f_j^k=\max _{l \in\{1, \cdots, k-1\}} a_{l, j}-a_{k, j}, \quad \forall j<k.
\end{equation}
The average forgetting of $k$-th tool is computed by normalizing against the number of tools seen previously:
\begin{equation}
\text{AF}_k=\frac{1}{k-1} \sum_{j=1}^{k-1} f_j^k.
\end{equation}
We report the average accuracy and average forgetting after the last tool learning.
\subsection{Comparisons with Video LLMs} \label{sec:5_2}
\begin{table*}[t]
	\centering
	\small
			\caption{\textbf{Comparisons (\%) with state-of-the-art methods on MVBench.} The tasks include Action Sequence (AS), Action Prediction (AP), Action Antonym (AA), Fine-grained Action (FA), Unexpected Action (UA), Object Existence (OE), Object Interaction (OI), Object Shuffle (OS), Moving Direction (MD), Action Localization (AL), Scene Transition (ST), Action Count (AC), Moving Count (MC), Moving Attribute (MA), State Change (SC), Fine-grained Pose (FP), Character Order (CO), Egocentric Navigation (EN), Episodic Reasoning (ER), Counterfactual Inference (CI), and the average of all 20 metrics (AVG). The best performance is in \textbf{bold} and the second best is \underline{underlined}.}
            \vspace{-2mm}
			\label{tab:mvbench}
			\renewcommand\arraystretch{1.1}
		\resizebox{\linewidth}{!}{
		\begin{tabular}{l|ccccccccccc}
			\toprule
			\textbf{Model}  & \textbf{AVG} & \textbf{AS} & \textbf{AP} & \textbf{AA} & \textbf{FA} & \textbf{UA} & \textbf{OE} & \textbf{OI} & \textbf{OS} & \textbf{MD} & \textbf{AL} \\
			\midrule
			InstructBLIP \cite{dai2024instructblip} & 32.5 & 20.0 & 16.5 & 46.0 & 24.5 & 46.0 & 51.0 & 26.0 & 37.5 & 22.0  & 23.0 \\
			LLaVA \cite{liu2024visual} & {36.0} & 28.0 & 39.5 & 63.0 & 30.5 & 39.0 & 53.0 & 41.0 & \underline{41.5} & 23.0 & 20.5 \\
			VideoChatGPT \cite{maaz2023video}  & {32.7} & 23.5 & 26.0 & 62.0 & 22.5 & 26.5 & 54.0 & 28.0 & 40.0 & 23.0  & 20.0 \\
			VideoLLaMA \cite{zhang2023video} & {34.1} & 27.5 & 25.5 & 51.0 & 29.0 & 39.0 & 48.0 & 40.5 & 38.0 & 22.5 & 22.5 \\
			VideoChat  \cite{li2023videochat} & {35.5} & 33.5 & 26.5 & 56.0 & 33.5 & 40.5 & 53.0 & 40.5 & 30.0 & 25.5 & 27.0 \\
			{VideoChat2$_\mathbf{text}$}  \cite{li2023mvbench}& {34.7} & 24.5 & 27.0 & 49.5 & 27.0 & 38.0 & 53.0 & 28.0 & 40.0 & 25.5 & 27.0 \\
			{VideoChat2}  \cite{li2023mvbench} & {{51.1}} & \textbf{66.0} & 47.5 & \textbf{83.5} & \textbf{49.5} & \underline{60.0} & 58.0 & \textbf{71.5} & \textbf{42.5} & 23.0 & 23.0 \\
			GPT-4V  \cite{gpt4v}  & {43.5} & 55.5 & \underline{63.5} & 72.0 & \underline{46.5} & \textbf{73.5} & 18.5 & 59.0 & 29.5 & 12.0 & \textbf{40.5} \\
                \rowcolor{customblue!20} \shortname$_\text{joint}$ (Ours) & {\textbf{53.4}} & \underline{64.0} & \textbf{65.0} & \underline{77.0} & 44.5 & 54.5 & \textbf{75.5} & \underline{70.0} & 34.0 & \textbf{36.5} & \underline{33.0} \\
                \rowcolor{customblue!20} \shortname$_\text{5$\times$2}$ (Ours) & {\underline{51.8}} & 62.0 & 61.0 & 75.0 & 42.5 & 52.5 & \underline{74.5} & 69.0 & 33.5 & \underline{35.0} & {32.0} \\
                \rowcolor{customblue!20} \shortname$_\text{10$\times$1}$ (Ours) & {50.6} & 60.0 & 60.5 & 73.5 & 42.0 & 51.5 & 73.5 &  68.5 & 32.5 & 33.0 & 30.0 \\
			\midrule
			\textbf{Model}  & \textbf{AVG} & \textbf{ST} & \textbf{AC} & \textbf{MC} & \textbf{MA} & \textbf{SC} & \textbf{FP} & \textbf{CO} & \textbf{EN} & \textbf{ER} & \textbf{CI} \\
			\midrule
			InstructBLIP \cite{dai2024instructblip} & {32.5} & 46.5 & \textbf{42.5} & 26.5 & 40.5 & 32.0 & 25.5 & 30.0 & 25.5 & 30.5 & 38.0 \\
			LLaVA \cite{liu2024visual} & {36.0}  & 45.0 & 34.0 & 20.5 & 38.5 & \underline{47.0} & 25.0 & 36.0 & 27.0 & 26.5 & 42.0 \\
			VideoChatGPT \cite{maaz2023video}  & {32.7} & 31.0 & 30.5 & 25.5 & 39.5 & \textbf{48.5} & 29.0 & 33.0 & 29.5 & 26.0 & 35.5 \\
			VideoLLaMA \cite{zhang2023video} & {34.1} & 43.0 & 34.0 & 22.5 & 32.5 & 45.5 & 32.5 & 40.0 & 30.0 & 21.0 & 37.0 \\
			VideoChat \cite{li2023videochat} & {35.5} & 48.5 & 35.0 & 20.5 & 42.5 & 46.0 & 26.5 & 41.0 & 23.5 & 23.5 & 36.0 \\
			{VideoChat2$_\mathbf{text}$} \cite{li2023mvbench}& {34.7} & 38.5 & \underline{41.5} & 27.5 & 32.5 & 46.5 & 26.5 & 36.0 & 33.0 & 32.0 & 40.0 \\
			{VideoChat2} \cite{li2023mvbench} & {51.1} & \textbf{88.5} & 39.0 & 42.0 & 58.5 & 44.0 & \textbf{49.0} & 36.5 & \textbf{35.0} & 40.5 & \textbf{65.5} \\
			GPT-4V \cite{gpt4v}  & {43.5} & 83.5 & 39.0 & 12.0 & 22.5 & 45.0 & \underline{47.5} & \textbf{52.0} & 31.0 & \textbf{59.0} & 11.0 \\
                \rowcolor{customblue!20} \shortname$_\text{joint}$ (Ours)  & {\textbf{53.4}} & \underline{88.0} & 37.5 & \textbf{53.5} & \textbf{75.5} & 35.5 & 42.5 & \underline{48.0} & \underline{34.5} & \underline{42.0} & \underline{57.5} \\
                \rowcolor{customblue!20} \shortname$_\text{5$\times$2}$ (Ours) & {\underline{51.8}} & 87.0 & 37.0 & \underline{52.5} & \underline{72.0} & 33.0 & 40.5 & \underline{48.0} & 33.5 & 40.5 & 55.0\\
                \rowcolor{customblue!20} \shortname$_\text{10$\times$1}$ (Ours) & {50.6} & 86.0 & 36.0 & 51.5 & 71.5 & 31.5 & 39.5 & 47.5 & 32.0 & 38.0 & 54.0  \\
			\bottomrule
		\end{tabular}
	}
\end{table*}

\begin{table*}[t]
\centering
\small
\caption{\textbf{Ablations studies on MVBench.} (a) training strategies; (b) training losses; (c) prompt selection mechanisms: T-based and V-based denote tool prompt selection based on text and visual features, respectively; (d) prompt positions. }
\vspace{-0.5em}
\renewcommand\arraystretch{1.1}
    \begin{subtable}[h]{0.24\textwidth}
        \begin{tabular}{ccx{25}}
	    \toprule
		\textbf{Stage 2} & \textbf{Stage 3}  & \textbf{AVG} \\
		\midrule
		\ding{55} & \ding{52} & 50.4\\
		\ding{52} & \ding{55} & 49.2\\		
		\ding{52} & \ding{52} & \textbf{51.8} \\		
		\bottomrule
	\end{tabular}
     \caption{}
     \label{tab:ablateTrain}
     \end{subtable}  \hfill
    \begin{subtable}[h]{0.24\textwidth}
        \begin{tabular}{x{25}x{25}x{25}}
		\toprule
		$\mathcal{L}_{q}$ & $\mathcal{L}_{c}$ & \textbf{AVG} \\
            \midrule
            \ding{55} & \ding{52} & 37.7\\
            \ding{52} & \ding{55} & 40.3 \\
            \ding{52} & \ding{52} & \textbf{51.8} \\
    	\bottomrule
	\end{tabular} 
    \caption{}
    \label{tab:ablateLoss}
    \end{subtable} \hfill
    \begin{subtable}[h]{0.25\textwidth}
        \begin{tabular}{ccc}
		\toprule
		\textbf{T-based} & \textbf{V-based} & \textbf{AVG} \\
            \midrule
            \ding{55} & \ding{52} & 45.3\\
            \ding{52} & \ding{55} & \textbf{51.8}\\
            \ding{52} & \ding{52} & 46.4 \\
    	\bottomrule
	\end{tabular} 
    \caption{}
    \label{tab:ablateSelect}
    \end{subtable} \hfill
     \begin{subtable}[h]{0.24\textwidth}
        \begin{tabular}{cx{30}}
		\toprule
		\textbf{Position}  & \textbf{AVG} \\
            \midrule
            \emph{tool-vision-text} &  51.7 \\
            \emph{vision-tool-text} &  51.7 \\		
            \emph{vision-text-tool} &  \textbf{51.8} \\	
    	\bottomrule
	\end{tabular} 
      \caption{}
      \label{tab:ablatePosition}
     \end{subtable}
\vspace{-1em}
\end{table*}
\begin{table*}[t]
    \small
    \setlength{\abovecaptionskip}{2pt} 
    \setlength{\belowcaptionskip}{-0.1cm}
    \renewcommand\arraystretch{1.1}
    \begin{center}
        \begin{subtable}[t]{0.48\linewidth}
            \centering
            \caption{\textbf{Comparisons with SOTA methods on test split of VideoTool.} 
            AR, VRD, and ASR denote action recognition, video relation detection and automatic speech recognition, respectively. 
            "Acc." denotes accuracy (\%) and "Score" denotes the GPT-evaluated score.}
            \vskip 10pt  
            \label{tab:toolEval}
            \resizebox{\linewidth}{!}{
            \begin{tabular}{lcccccccc}
                \toprule
                \multirow{2}{*}{\textbf{Method}}  & \multicolumn{2}{c}{\textbf{AR}} & \multicolumn{2}{c}{\textbf{VRD}} & \multicolumn{2}{c}{\textbf{ASR}}  \\
                \cmidrule(lr){2-3} \cmidrule(lr){4-5} \cmidrule(l){6-7} 
                & \textbf{Acc.} & \textbf{Score} & \textbf{Acc.} & \textbf{Score} & \textbf{Acc.} & \textbf{Score} \\
                \midrule
                VideoLLaVA   & 47.7 & 2.89 & 6.0 & 0.82 & 1.83 & 0.77  \\
                VideoChat2   & 51.4 & 3.13 & 14.0 & 1.36 & 3.60 & 0.87  \\
                VideoChat   &  29.0 & 2.31 & 4.5 & 0.90 & 2.80 & 1.28   \\
                Videoagent   &  68.13 & 3.16 & 16.7 & 1.29 & 1.79 & 0.83   \\
                VITAL   &  52.23 & 3.09 & 15.28 & 1.24 & 1.98 & 0.85   \\
                \rowcolor{customblue!20} \shortname$_\text{joint}$  & \textbf{77.9} & \textbf{3.84} & \textbf{24.0} & \textbf{1.75} & \textbf{24.2} & \textbf{1.95}  \\
                \rowcolor{customblue!20} \shortname$_\text{5$\times$2}$ & 74.8 & 3.60 & 21.3 & 1.48 & 22.6 & 1.50  \\
                \rowcolor{customblue!20} \shortname$_\text{10$\times$1}$  & 73.5 & 3.48 & 19.6 & 1.27 & 21.9 & 1.48  \\
                \addlinespace[6pt] 
                \bottomrule
            \end{tabular}}
        \end{subtable}
        \hfill
        \begin{subtable}[t]{0.48\linewidth}
            \centering
            \caption{\textbf{Comparisons (\%) with continual learning methods on VideoTool.} 
            AA and AF denote average accuracy and average forgetting, respectively.}
            \vskip 27pt  
            \label{tab:CLbench}
            \resizebox{\linewidth}{!}{
            \begin{tabular}{lcccc}
                \toprule
                \multirow{2}{*}{\textbf{Method}} & \multicolumn{2}{c}{\textbf{Five Groups}} & \multicolumn{2}{c}{\textbf{Ten Groups}} \\
                \cmidrule(lr){2-3} \cmidrule(lr){4-5} 
                & \textbf{AA↑} & \textbf{AF↓} & \textbf{AA↑} & \textbf{AF↓} \\
                \midrule
                Sequential  & 48.6 & 35.3  & 42.3 & 39.7 \\
                Rehearsal (10/Tool) & 52.9 & 32.1  &  48.2 & 36.3  \\
                Rehearsal (30/Tool) & 55.2 & 30.2  & 49.3 & 35.5\\
                Rehearsal (50/Tool) & 59.3 & 27.4 & 50.2 & 32.7\\
                L2P \cite{wang2022learning} & 72.4 & 8.3 & 65.7 & 12.4\\
                Dual \cite{wang2022dualprompt} & 75.1 & 6.9 & 68.6 & 7.5   \\
                CODA \cite{smith2023coda} & 76.3 & 7.0  & 71.0 & 7.2 \\
                \rowcolor{customblue!20} \shortname (Ours) & \textbf{79.8} & \textbf{5.8} & \textbf{74.7} & \textbf{6.4}  \\				
                \bottomrule
            \end{tabular}}
        \end{subtable}
    \end{center}
\end{table*}



We set up three model variants: \textbf{1)} COLT$_\text{5$\times$2}$ receives five successive groups of data and each group contains data of two tools; \textbf{2)} COLT$_\text{10$\times$1}$ is defined similarly; \textbf{3)} COLT$_\text{joint}$ receives all tool data at once and is trained with all the data collectively. The performance of COLT$_\text{joint}$ is regarded as the upper-bound of the continual learning counterpart. 

\textbf{Evaluation Benchmarks.} Our experiments are carried out on both established video LLM benchmarks and self-built tool-using datasets: 1) \textbf{zero-shot video-question answering}. We experiment on commonly used open-ended question-answer datasets: MSVD-QA \cite{chen2011collecting}, MSRVTT-QA \cite{xu2016msr}, and ActivityNet-QA \cite{yu2019activitynet}. Following \cite{li2023videochat,lin2023video}, we use GPT-assisted evaluation to assess the model’s capabilities by reporting the accuracy and relative score; 2) \textbf{MVBench} \cite{li2023mvbench}. This benchmark consists of 20 demanding video tasks, each comprising 200 samples presented as multiple-choice questions. These tasks offer a thorough and unbiased evaluation of a model's capacity to comprehend videos. We report the choice accuracy as the metric; 
3) \textbf{VideoTool}. We built this dataset to probe the abilities enabled by tool proficiency. Since most existing video LLMs only support textual outputs, we select three tools (\ie, action recognition, video relation detection, and automatic speech recognition) and compare our COLT to state-of-the-art video LLMs \cite{lin2023video,li2023mvbench,li2023videochat}. The GPT-evaluated accuracy and scores are reported. 


\textbf{Performance Analysis.} The comparison results on zero-shot video-question answering, MVBench, and VideoTool test split are summarized in Table~\ref{tab:zeroqa}, Table~\ref{tab:mvbench}, and Table~\ref{tab:toolEval}, respectively. We can conclude with the following findings. \textbf{1)} Both \shortname$_\text{joint}$ and \shortname$_\text{5$\times$2}$ demonstrate superior performance compared to preceding state-of-the-art methodologies. For example on MSRVTT-QA (\cf Table~\ref{tab:zeroqa}), \shortname$_\text{joint}$ remarkably surpasses the previous best performing method VideoChat2 \cite{li2023mvbench} by 8.2\% on the metric of accuracy; \textbf{2)} On the test set of VideoToolBench (\cf Table~\ref{tab:toolEval}), our COLT consistently outperforms prior approaches by a clear margin, including both existing video-VQA models and representative tool-based agents, demonstrating its effectiveness in tool-intensive scenarios; \textbf{3)} Even with a rather lengthy learning curve, \shortname$_\text{10$\times$1}$ still achieves comparable performance with previous SOTA methods; \textbf{4)} The performance of \shortname$_\text{10$\times$1}$ is slightly worse than \shortname$_\text{5$\times$2}$, which is consistent with the intuition that tool learning becomes increasingly challenging when facing a lengthening tool-chain. 
\subsection{Comparisons with Continual Learning Methods} \label{sec:5_3}
We additionally compare the proposed \shortname with popular continual learning methods on VideoTool test split to demonstrate the life-long tool-usage learning capability of our method.

\textbf{Evaluation Metrics.} To evaluate how the system retains tool-using knowledge over time, we adapt the conventional continual learning metric \cite{wang2024comprehensive,chaudhry2018riemannian} to our tool-using scenario. We set up two metrics: 1) \emph{average accuracy} of a specific tool denotes the overall call accuracy during the incremental learning process; 2) \emph{average forgetting} is defined as the mean difference between the maximum tool call accuracy throughout the past learning process and the current tool call accuracy. A reduction in the value of average forgetting is indicative of enhanced continual learning capability. Refer to the supplementary materials for the formula and detailed explanations.
\textbf{Compared Methods.} We set comparison experiments as follows: 1) \emph{Sequential training}: training on new data with the pre-trained weights on previous tool data as initialization; 2) \emph{Rehearsal training}: replay past tool data (\ie, ``buffer'') and combine them with new data. Buffer size is respectively set to 10/30/50 for each tool in experiments; 3) Popular continual learning methods including L2P \cite{wang2022learning}, Dual \cite{wang2022dualprompt}, and CODA \cite{smith2023coda}. 

\textbf{Performance Analysis.}
We conduct experiments in two settings, \ie, five/ten groups with two/one tools per group. 
As shown in Table~\ref{tab:CLbench}, COLT consistently exhibits superior performance in both 
average accuracy and forgetting across both settings. 
For example, under the five-group setting, COLT outperforms CODA~\cite{smith2023coda} by 3.5\% in average accuracy, 
highlighting its effectiveness in mitigating catastrophic forgetting. 
Notably, despite being given rehearsal buffers of size 10/30/50 per tool, rehearsal-based methods remain 
significantly weaker (Table~\ref{tab:CLbench}), as they depend on storing and replaying past data and still 
struggle with interference across heterogeneous tools. 
In contrast, COLT achieves stronger retention \emph{without} accessing old samples, demonstrating that its 
codebook-based tool memory provides a more stable and efficient continual-learning mechanism than rehearsal.
\subsection{Ablation Studies} \label{sec:5_4}

We conduct extensive ablation studies to provide more insights into our proposed COLT. The experiments are conducted on MVBench~\cite{li2023mvbench} with the continual learning model variant \shortname$_\text{5$\times$2}$.




\textbf{Ablation on Training Strategies.} Our COLT adopts a three-stage training pipeline. We conduct ablation studies on the training process by respectively skipping the second and third stages of training. The comparison results on MVBench \cite{li2023mvbench} in Table \ref{tab:ablateTrain} demonstrate that both stages are crucial to the final performance, \eg, skipping the second stage leads to a 1.4\% drop in average scores.

\textbf{Ablation on Training Loss.} We use the straight-through estimator \cite{van2017neural,bengio2013estimating} for the training of stage 2 and stage 3 (\cf Eq.~\eqref{eq:lossstage2}), which contains the quantisation loss $\mathcal{L}_{q}$ and commitment loss $\mathcal{L}_{c}$ to enables the mutual updates between the selected tool prompts and query features. The ablation results in Table \ref{tab:ablateLoss} underscore the significance of both quantisation loss $\mathcal{L}_{q}$ and commitment loss $\mathcal{L}_{c}$.

\textbf{Ablation on Prompt Selection.} Recall that we select the most matched tool prompts based on the similarity between the textual instruction features and codebook (\ie, \emph{text}-\emph{codebook}). Here we ablate on the prompt selection based on the \emph{vision}-\emph{codebook} similarity or the average similarities of both. The results are listed in Table \ref{tab:ablateSelect}, which showcases that the \emph{text}-\emph{codebook} similarity is more reliable. That could be attributed to the fact it is easier for video LLMs to decide whether to invoke or which specific tool to invoke from the user instructions instead of the input visual data.

\textbf{Ablation on Tool Prompt Position.} We ablate on the insertion position of the tool prompt. We've listed three options, including \emph{tool-vision-text}, \emph{vision-tool-text}, and \emph{vision-text-tool}, demonstrating placing the tool feature before/between/after the vision and text features, respectively. The findings in Table \ref{tab:ablatePosition} indicate that the performance is insensitive with regard to the tool prompt position.


\newcommand{\cellshrink}[2][0.9]{%
  \begingroup
    \sbox0{#2}%
    \makebox[\wd0][c]{\scalebox{#1}{#2}}%
  \endgroup
}

\begin{figure*}[t]
  \centering
  \small

  \newlength{\panelheight}
  \setlength{\panelheight}{2.8in} 

  \begin{subfigure}[t]{0.48\linewidth}
    \begin{minipage}[t][\panelheight][t]{\linewidth}
      \vspace*{0pt} 
      \centering
      \includegraphics[width=\linewidth,height=\panelheight,keepaspectratio]{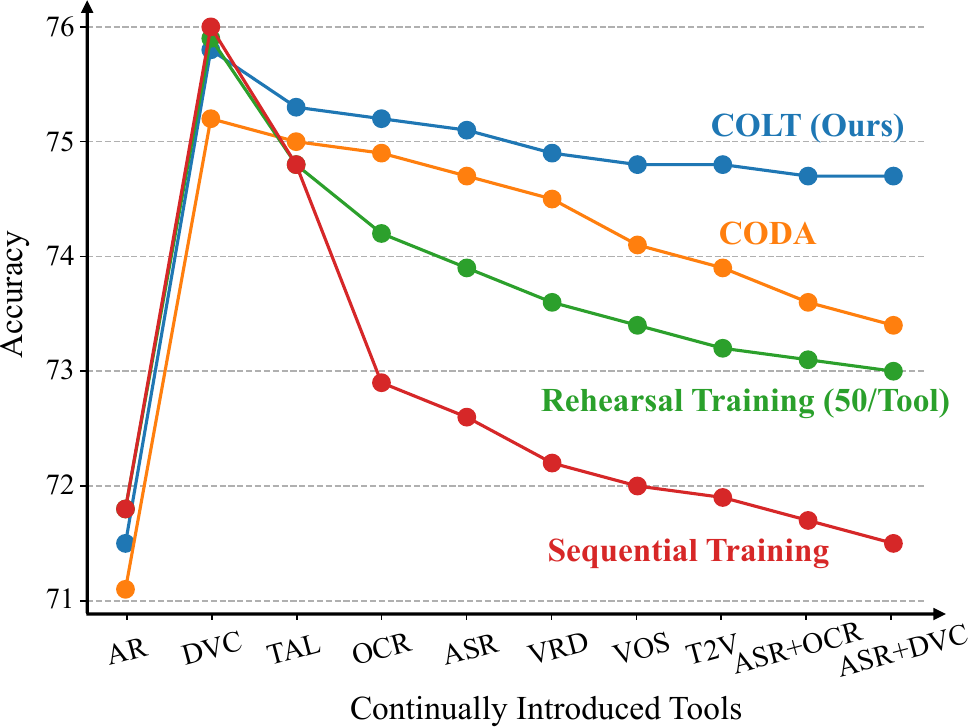}
      \vspace*{-10pt}
      \caption{\textbf{Accuracy of zero-shot video-question answering on MSVD \vs continually introduced tasks.}
      The specific tool name for each abbreviation is available in Table~\ref{table:toollist}.}
      \label{fig:acc_vs_tool}
    \end{minipage}
  \end{subfigure}
  \hfill
  \begin{subfigure}[t]{0.48\linewidth}
    \begin{minipage}[t][\panelheight][t]{\linewidth}
      \vspace*{6pt} 
      \centering
      \renewcommand{\arraystretch}{1.3}
      \setlength{\extrarowheight}{1pt}

      {\footnotesize
      \resizebox{\linewidth}{!}{%
        \begin{tabular}{cccccc}
          \toprule
          \cellshrink{$N$} & \cellshrink{20} & \cellshrink{30} & \cellshrink{40} & \cellshrink{50} & \cellshrink{60} \\
          \midrule
          \cellshrink{AVG} & \cellshrink{48.4} & \cellshrink{50.9} & \cellshrink{51.4} & \textbf{\cellshrink{51.8}} & \cellshrink{51.8}\\
          \midrule \addlinespace[25pt]  \midrule
          \cellshrink{$K$} & \cellshrink{1} & \cellshrink{2} & \cellshrink{3} & \cellshrink{4} & \cellshrink{5} \\
          \midrule
          \cellshrink{AVG} & \cellshrink{50.8}  & \cellshrink{51.6} & \textbf{\cellshrink{51.8}} & \cellshrink{51.3} & \cellshrink{51.3} \\
          \bottomrule
        \end{tabular}%
      }}
      \vspace*{27pt}
      \caption{\textbf{Ablations on hyper-parameters} including the codebook size $N$ and the number of selected prompt $K$.}
      \label{tab:ablaParam}
    \end{minipage}
  \end{subfigure}
\end{figure*}

\textbf{Ablation on hyper-parameters.} We conduct ablation studies on the codebook size $N$ and the number of selected prompts $K$. The comparison results are listed in Figure \ref{tab:ablaParam}. As shown, the average performance is positively correlated with the codebook size $N$ and reaches saturation at $T=50$. The optimal results are achieved when setting $K=3$.

\textbf{Visualizations of performance \vs continually introduced tasks.} To intuitively show the impact of incrementally introducing tools, we show the phased zero-shot video-question answering performance under different training strategies. Specifically, we report the accuracy score on MSVD-QA dataset in Figure \ref{fig:acc_vs_tool}. As shown, equipping video LLMs with dense video caption tools leads to the significant performance boost for all four training strategies. Besides, our proposed COLT shows better learning \emph{stability}, \ie, less forgetfulness when faced with new tool data.

\section{Limitations}\label{sec:limitation}

While COLT demonstrates strong continual tool-use capabilities, several limitations remain. First, VideoTool is primarily GPT-generated, which may introduce distribution biases and limit the dataset’s real-world diversity compared with fully human-curated corpora. Second, the current dataset focuses mainly on single-tool and simple multi-tool compositions, which do not fully capture the complexity, interdependence, and noise present in real-world tool ecosystems, where tool chains may be substantially longer and more intricate. Future work includes extending VideoTool to more diverse and human-in-the-loop scenarios, scaling COLT to larger and more complex tool vocabularies, and exploring reinforcement learning or other adaptive strategies to support dynamic and automatic tool composition.

\section{Conclusions}\label{sec:conclusion}
In this work, we present COLT, which continually learns new tool-using knowledge in a data-stream scenario for general-purpose video understanding. To achieve this, we propose a learnable tool codebook where the specific tool is retrieved and activated according to the similarities with the user instructions. Due to the absence of a video-centric tool-using instruction-tuning dataset within the community, we devised VideoTool to address this deficiency and foster the exploration of tool-using capacities of video LLMs. Experimental results indicate that our proposed COLT adeptly invokes the necessary tools with precision, thereby achieving state-of-the-art performance on widely used benchmarks and the proposed VideoTool test split.

\subsubsection*{Broader Impact Statement}
The ability of our COLT to continually learn new tool usage without catastrophic forgetting may support the development of \emph{personalized} video LLMs. Data within specific domains or pertaining to particular users can be incrementally fine-tuned, leading to an ever-evolving and personalized intelligent assistant. However, there also exists the risk that the model could be exploited for malicious purposes. Besides, it may raise privacy concerns, especially if it is deployed in surveillance systems or social media platforms. 

\bibliography{main}

@misc{zhang2025thinkingvideosmultimodaltoolaugmented,
      title={Thinking With Videos: Multimodal Tool-Augmented Reinforcement Learning for Long Video Reasoning}, 
      author={Haoji Zhang and Xin Gu and Jiawen Li and Chixiang Ma and Sule Bai and Chubin Zhang and Bowen Zhang and Zhichao Zhou and Dongliang He and Yansong Tang},
      year={2025},
      eprint={2508.04416},
      archivePrefix={arXiv},
      primaryClass={cs.CV},
      url={https://arxiv.org/abs/2508.04416}, 
}

@article{achiam2023gpt,
	title={Gpt-4 technical report},
	author={Achiam, Josh and Adler, Steven and Agarwal, Sandhini and Ahmad, Lama and Akkaya, Ilge and Aleman, Florencia Leoni and Almeida, Diogo and Altenschmidt, Janko and Altman, Sam and Anadkat, Shyamal and others},
	journal={arXiv preprint arXiv:2303.08774},
	year={2023}
}

@article{brown2020language,
	title={Language models are few-shot learners},
	author={Brown, Tom and Mann, Benjamin and Ryder, Nick and Subbiah, Melanie and Kaplan, Jared D and Dhariwal, Prafulla and Neelakantan, Arvind and Shyam, Pranav and Sastry, Girish and Askell, Amanda and others},
	journal={Advances in neural information processing systems},
	volume={33},
	pages={1877--1901},
	year={2020}
}

@article{anil2023palm,
	title={Palm 2 technical report},
	author={Anil, Rohan and Dai, Andrew M and Firat, Orhan and Johnson, Melvin and Lepikhin, Dmitry and Passos, Alexandre and Shakeri, Siamak and Taropa, Emanuel and Bailey, Paige and Chen, Zhifeng and others},
	journal={arXiv preprint arXiv:2305.10403},
	year={2023}
}

@article{touvron2023llama,
	title={Llama: Open and efficient foundation language models},
	author={Touvron, Hugo and Lavril, Thibaut and Izacard, Gautier and Martinet, Xavier and Lachaux, Marie-Anne and Lacroix, Timoth{\'e}e and Rozi{\`e}re, Baptiste and Goyal, Naman and Hambro, Eric and Azhar, Faisal and others},
	journal={arXiv preprint arXiv:2302.13971},
	year={2023}
}

@article{touvron2023llama2,
	title={Llama 2: Open foundation and fine-tuned chat models},
	author={Touvron, Hugo and Martin, Louis and Stone, Kevin and Albert, Peter and Almahairi, Amjad and Babaei, Yasmine and Bashlykov, Nikolay and Batra, Soumya and Bhargava, Prajjwal and Bhosale, Shruti and others},
	journal={arXiv preprint arXiv:2307.09288},
	year={2023}
}

@inproceedings{li2023blip,
	title={Blip-2: Bootstrapping language-image pre-training with frozen image encoders and large language models},
	author={Li, Junnan and Li, Dongxu and Savarese, Silvio and Hoi, Steven},
	booktitle={International conference on machine learning},
	pages={19730--19742},
	year={2023},
	organization={PMLR}
}

@article{alayrac2022flamingo,
	title={Flamingo: a visual language model for few-shot learning},
	author={Alayrac, Jean-Baptiste and Donahue, Jeff and Luc, Pauline and Miech, Antoine and Barr, Iain and Hasson, Yana and Lenc, Karel and Mensch, Arthur and Millican, Katherine and Reynolds, Malcolm and others},
	journal={Advances in neural information processing systems},
	volume={35},
	pages={23716--23736},
	year={2022}
}

@inproceedings{zhu2023minigpt,
	title={MiniGPT-4: Enhancing Vision-Language Understanding with Advanced Large Language Models},
	author={Zhu, Deyao and Chen, Jun and Shen, Xiaoqian and Li, Xiang and Elhoseiny, Mohamed},
	booktitle={The Twelfth International Conference on Learning Representations},
	year={2023}
}

@article{liu2024visual,
	title={Visual instruction tuning},
	author={Liu, Haotian and Li, Chunyuan and Wu, Qingyang and Lee, Yong Jae},
	journal={Advances in neural information processing systems},
	volume={36},
	year={2024}
}

@article{wu2023next,
	title={Next-gpt: Any-to-any multimodal llm},
	author={Wu, Shengqiong and Fei, Hao and Qu, Leigang and Ji, Wei and Chua, Tat-Seng},
	journal={arXiv preprint arXiv:2309.05519},
	year={2023}
}

@article{zhan2024anygpt,
	title={AnyGPT: Unified Multimodal LLM with Discrete Sequence Modeling},
	author={Zhan, Jun and Dai, Junqi and Ye, Jiasheng and Zhou, Yunhua and Zhang, Dong and Liu, Zhigeng and Zhang, Xin and Yuan, Ruibin and Zhang, Ge and Li, Linyang and others},
	journal={arXiv preprint arXiv:2402.12226},
	year={2024}
}

@article{lin2023video,
	title={Video-llava: Learning united visual representation by alignment before projection},
	author={Lin, Bin and Zhu, Bin and Ye, Yang and Ning, Munan and Jin, Peng and Yuan, Li},
	journal={arXiv preprint arXiv:2311.10122},
	year={2023}
}

@inproceedings{zhang2023video,
	title={Video-LLaMA: An Instruction-tuned Audio-Visual Language Model for Video Understanding},
	author={Zhang, Hang and Li, Xin and Bing, Lidong},
	booktitle={Proceedings of the 2023 Conference on Empirical Methods in Natural Language Processing: System Demonstrations},
	pages={543--553},
	year={2023}
}

@article{li2023videochat,
	title={Videochat: Chat-centric video understanding},
	author={Li, KunChang and He, Yinan and Wang, Yi and Li, Yizhuo and Wang, Wenhai and Luo, Ping and Wang, Yali and Wang, Limin and Qiao, Yu},
	journal={arXiv preprint arXiv:2305.06355},
	year={2023}
}

@article{maaz2023video,
	title={Video-chatgpt: Towards detailed video understanding via large vision and language models},
	author={Maaz, Muhammad and Rasheed, Hanoona and Khan, Salman and Khan, Fahad Shahbaz},
	journal={arXiv preprint arXiv:2306.05424},
	year={2023}
}

@article{li2023llama,
	title={LLaMA-VID: An image is worth 2 tokens in large language models},
	author={Li, Yanwei and Wang, Chengyao and Jia, Jiaya},
	journal={arXiv preprint arXiv:2311.17043},
	year={2023}
}

@article{jin2024video,
	title={Video-LaVIT: Unified Video-Language Pre-training with Decoupled Visual-Motional Tokenization},
	author={Jin, Yang and Sun, Zhicheng and Xu, Kun and Chen, Liwei and Jiang, Hao and Huang, Quzhe and Song, Chengru and Liu, Yuliang and Zhang, Di and Song, Yang and others},
	journal={arXiv preprint arXiv:2402.03161},
	year={2024}
}

@article{munasinghe2023pg,
	title={Pg-video-llava: Pixel grounding large video-language models},
	author={Munasinghe, Shehan and Thushara, Rusiru and Maaz, Muhammad and Rasheed, Hanoona Abdul and Khan, Salman and Shah, Mubarak and Khan, Fahad},
	journal={arXiv preprint arXiv:2311.13435},
	year={2023}
}

@article{liu2024st,
	title={ST-LLM: Large Language Models Are Effective Temporal Learners},
	author={Liu, Ruyang and Li, Chen and Tang, Haoran and Ge, Yixiao and Shan, Ying and Li, Ge},
	journal={arXiv preprint arXiv:2404.00308},
	year={2024}
}

@article{jin2023chat,
	title={Chat-univi: Unified visual representation empowers large language models with image and video understanding},
	author={Jin, Peng and Takanobu, Ryuichi and Zhang, Caiwan and Cao, Xiaochun and Yuan, Li},
	journal={arXiv preprint arXiv:2311.08046},
	year={2023}
}

@article{liu2023one,
	title={One for all: Video conversation is feasible without video instruction tuning},
	author={Liu, Ruyang and Li, Chen and Ge, Yixiao and Shan, Ying and Li, Thomas H and Li, Ge},
	journal={arXiv preprint arXiv:2309.15785},
	year={2023}
}

@article{gao2023assistgpt,
	title={Assistgpt: A general multi-modal assistant that can plan, execute, inspect, and learn},
	author={Gao, Difei and Ji, Lei and Zhou, Luowei and Lin, Kevin Qinghong and Chen, Joya and Fan, Zihan and Shou, Mike Zheng},
	journal={arXiv preprint arXiv:2306.08640},
	year={2023}
}

@article{qin2023toolllm,
	title={Toolllm: Facilitating large language models to master 16000+ real-world apis},
	author={Qin, Yujia and Liang, Shihao and Ye, Yining and Zhu, Kunlun and Yan, Lan and Lu, Yaxi and Lin, Yankai and Cong, Xin and Tang, Xiangru and Qian, Bill and others},
	journal={arXiv preprint arXiv:2307.16789},
	year={2023}
}

@article{yang2024doraemongpt,
	title={Doraemongpt: Toward understanding dynamic scenes with large language models},
	author={Yang, Zongxin and Chen, Guikun and Li, Xiaodi and Wang, Wenguan and Yang, Yi},
	journal={arXiv preprint arXiv:2401.08392},
	year={2024}
}

@article{fan2024videoagent,
	title={VideoAgent: A Memory-augmented Multimodal Agent for Video Understanding},
	author={Fan, Yue and Ma, Xiaojian and Wu, Rujie and Du, Yuntao and Li, Jiaqi and Gao, Zhi and Li, Qing},
	journal={arXiv preprint arXiv:2403.11481},
	year={2024}
}

@article{wang2024videoagent,
	title={VideoAgent: Long-form Video Understanding with Large Language Model as Agent},
	author={Wang, Xiaohan and Zhang, Yuhui and Zohar, Orr and Yeung-Levy, Serena},
	journal={arXiv preprint arXiv:2403.10517},
	year={2024}
}

@misc{EasyOCR,
	author = {JaidedAI},
	title = {EasyOCR},
	year = {2023},
	howpublished = {\url{https://github.com/JaidedAI/EasyOCR}},
}

@article{wang2022internvideo,
	title={Internvideo: General video foundation models via generative and discriminative learning},
	author={Wang, Yi and Li, Kunchang and Li, Yizhuo and He, Yinan and Huang, Bingkun and Zhao, Zhiyu and Zhang, Hongjie and Xu, Jilan and Liu, Yi and Wang, Zun and others},
	journal={arXiv preprint arXiv:2212.03191},
	year={2022}
}

@inproceedings{wang2021end,
	title={End-to-end video instance segmentation with transformers},
	author={Wang, Yuqing and Xu, Zhaoliang and Wang, Xinlong and Shen, Chunhua and Cheng, Baoshan and Shen, Hao and Xia, Huaxia},
	booktitle={Proceedings of the IEEE/CVF conference on computer vision and pattern recognition},
	pages={8741--8750},
	year={2021}
}

@inproceedings{radford2023robust,
	title={Robust speech recognition via large-scale weak supervision},
	author={Radford, Alec and Kim, Jong Wook and Xu, Tao and Brockman, Greg and McLeavey, Christine and Sutskever, Ilya},
	booktitle={International Conference on Machine Learning},
	pages={28492--28518},
	year={2023},
	organization={PMLR}
}

@inproceedings{shang2017video,
	title={Video visual relation detection},
	author={Shang, Xindi and Ren, Tongwei and Guo, Jingfan and Zhang, Hanwang and Chua, Tat-Seng},
	booktitle={Proceedings of the 25th ACM international conference on Multimedia},
	pages={1300--1308},
	year={2017}
}

@article{yang2023mm,
	title={Mm-react: Prompting chatgpt for multimodal reasoning and action},
	author={Yang, Zhengyuan and Li, Linjie and Wang, Jianfeng and Lin, Kevin and Azarnasab, Ehsan and Ahmed, Faisal and Liu, Zicheng and Liu, Ce and Zeng, Michael and Wang, Lijuan},
	journal={arXiv preprint arXiv:2303.11381},
	year={2023}
}

@inproceedings{yao2022react,
	title={ReAct: Synergizing Reasoning and Acting in Language Models},
	author={Yao, Shunyu and Zhao, Jeffrey and Yu, Dian and Du, Nan and Shafran, Izhak and Narasimhan, Karthik R and Cao, Yuan},
	booktitle={The Eleventh International Conference on Learning Representations},
	year={2022}
}

@article{zhang2023llama,
	title={Llama-adapter: Efficient fine-tuning of language models with zero-init attention},
	author={Zhang, Renrui and Han, Jiaming and Liu, Chris and Gao, Peng and Zhou, Aojun and Hu, Xiangfei and Yan, Shilin and Lu, Pan and Li, Hongsheng and Qiao, Yu},
	journal={arXiv preprint arXiv:2303.16199},
	year={2023}
}

@article{li2023mvbench,
	title={Mvbench: A comprehensive multi-modal video understanding benchmark},
	author={Li, Kunchang and Wang, Yali and He, Yinan and Li, Yizhuo and Wang, Yi and Liu, Yi and Wang, Zun and Xu, Jilan and Chen, Guo and Luo, Ping and others},
	journal={arXiv preprint arXiv:2311.17005},
	year={2023}
}

@inproceedings{zhu2023languagebind,
	title={LanguageBind: Extending Video-Language Pretraining to N-modality by Language-based Semantic Alignment},
	author={Zhu, Bin and Lin, Bin and Ning, Munan and Yan, Yang and Cui, Jiaxi and HongFa, WANG and Pang, Yatian and Jiang, Wenhao and Zhang, Junwu and Li, Zongwei and others},
	booktitle={The Twelfth International Conference on Learning Representations},
	year={2023}
}

@inproceedings{sennrich2016neural,
	title={Neural Machine Translation of Rare Words with Subword Units},
	author={Sennrich, Rico and Haddow, Barry and Birch, Alexandra},
	booktitle={Proceedings of the 54th Annual Meeting of the Association for Computational Linguistics (Volume 1: Long Papers)},
	pages={1715--1725},
	year={2016}
}

@article{schuhmann2021laion,
  title={Laion-400m: Open dataset of clip-filtered 400 million image-text pairs},
  author={Schuhmann, Christoph and Vencu, Richard and Beaumont, Romain and Kaczmarczyk, Robert and Mullis, Clayton and Katta, Aarush and Coombes, Theo and Jitsev, Jenia and Komatsuzaki, Aran},
  journal={arXiv preprint arXiv:2111.02114},
  year={2021}
}

@inproceedings{changpinyo2021conceptual,
	title={Conceptual 12m: Pushing web-scale image-text pre-training to recognize long-tail visual concepts},
	author={Changpinyo, Soravit and Sharma, Piyush and Ding, Nan and Soricut, Radu},
	booktitle={Proceedings of the IEEE/CVF conference on computer vision and pattern recognition},
	pages={3558--3568},
	year={2021}
}

@article{bengio2013estimating,
	title={Estimating or propagating gradients through stochastic neurons for conditional computation},
	author={Bengio, Yoshua and L{\'e}onard, Nicholas and Courville, Aaron},
	journal={arXiv preprint arXiv:1308.3432},
	year={2013}
}

@article{van2017neural,
	title={Neural discrete representation learning},
	author={Van Den Oord, Aaron and Vinyals, Oriol and others},
	journal={Advances in neural information processing systems},
	volume={30},
	year={2017}
}

@inproceedings{xu2016msr,
	title={Msr-vtt: A large video description dataset for bridging video and language},
	author={Xu, Jun and Mei, Tao and Yao, Ting and Rui, Yong},
	booktitle={Proceedings of the IEEE conference on computer vision and pattern recognition},
	pages={5288--5296},
	year={2016}
}

@inproceedings{chen2011collecting,
	title={Collecting highly parallel data for paraphrase evaluation},
	author={Chen, David and Dolan, William B},
	booktitle={Proceedings of the 49th annual meeting of the association for computational linguistics: human language technologies},
	pages={190--200},
	year={2011}
}

@inproceedings{yu2019activitynet,
	title={Activitynet-qa: A dataset for understanding complex web videos via question answering},
	author={Yu, Zhou and Xu, Dejing and Yu, Jun and Yu, Ting and Zhao, Zhou and Zhuang, Yueting and Tao, Dacheng},
	booktitle={Proceedings of the AAAI Conference on Artificial Intelligence},
	volume={33},
	pages={9127--9134},
	year={2019}
}

@inproceedings{liu2023improved,
	title={Improved Baselines with Visual Instruction Tuning},
	author={Liu, Haotian and Li, Chunyuan and Li, Yuheng and Lee, Yong Jae},
	booktitle={NeurIPS 2023 Workshop on Instruction Tuning and Instruction Following},
	year={2023}
}

@article{dai2024instructblip,
	title={Instructblip: Towards general-purpose vision-language models with instruction tuning},
	author={Dai, Wenliang and Li, Junnan and Li, Dongxu and Tiong, Anthony Meng Huat and Zhao, Junqi and Wang, Weisheng and Li, Boyang and Fung, Pascale N and Hoi, Steven},
	journal={Advances in Neural Information Processing Systems},
	volume={36},
	year={2024}
}

@misc{gpt4v,
	title={GPT-4V(ision) System Card},
	author={OpenAI},
	howpublished = {\url{https://api.semanticscholar.org/CorpusID:263218031}},
	year={2023}
}

@article{wang2024comprehensive,
  title={A comprehensive survey of continual learning: Theory, method and application},
  author={Wang, Liyuan and Zhang, Xingxing and Su, Hang and Zhu, Jun},
  journal={IEEE Transactions on Pattern Analysis and Machine Intelligence},
  year={2024},
  publisher={IEEE}
}

@inproceedings{chaudhry2018riemannian,
  title={Riemannian walk for incremental learning: Understanding forgetting and intransigence},
  author={Chaudhry, Arslan and Dokania, Puneet K and Ajanthan, Thalaiyasingam and Torr, Philip HS},
  booktitle={Proceedings of the European conference on computer vision (ECCV)},
  pages={532--547},
  year={2018}
}

@article{li2017learning,
  title={Learning without forgetting},
  author={Li, Zhizhong and Hoiem, Derek},
  journal={IEEE transactions on pattern analysis and machine intelligence},
  volume={40},
  number={12},
  pages={2935--2947},
  year={2017},
  publisher={IEEE}
}

@article{kirkpatrick2017overcoming,
  title={Overcoming catastrophic forgetting in neural networks},
  author={Kirkpatrick, James and Pascanu, Razvan and Rabinowitz, Neil and Veness, Joel and Desjardins, Guillaume and Rusu, Andrei A and Milan, Kieran and Quan, John and Ramalho, Tiago and Grabska-Barwinska, Agnieszka and others},
  journal={Proceedings of the national academy of sciences},
  volume={114},
  number={13},
  pages={3521--3526},
  year={2017},
  publisher={National Acad Sciences}
}

@inproceedings{wang2022learning,
  title={Learning to prompt for continual learning},
  author={Wang, Zifeng and Zhang, Zizhao and Lee, Chen-Yu and Zhang, Han and Sun, Ruoxi and Ren, Xiaoqi and Su, Guolong and Perot, Vincent and Dy, Jennifer and Pfister, Tomas},
  booktitle={Proceedings of the IEEE/CVF Conference on Computer Vision and Pattern Recognition},
  pages={139--149},
  year={2022}
}

@inproceedings{wang2022dualprompt,
  title={Dualprompt: Complementary prompting for rehearsal-free continual learning},
  author={Wang, Zifeng and Zhang, Zizhao and Ebrahimi, Sayna and Sun, Ruoxi and Zhang, Han and Lee, Chen-Yu and Ren, Xiaoqi and Su, Guolong and Perot, Vincent and Dy, Jennifer and others},
  booktitle={European Conference on Computer Vision},
  pages={631--648},
  year={2022},
  organization={Springer}
}

@inproceedings{smith2023coda,
  title={Coda-prompt: Continual decomposed attention-based prompting for rehearsal-free continual learning},
  author={Smith, James Seale and Karlinsky, Leonid and Gutta, Vyshnavi and Cascante-Bonilla, Paola and Kim, Donghyun and Arbelle, Assaf and Panda, Rameswar and Feris, Rogerio and Kira, Zsolt},
  booktitle={Proceedings of the IEEE/CVF Conference on Computer Vision and Pattern Recognition},
  pages={11909--11919},
  year={2023}
}

@article{bonicelli2022effectiveness,
  title={On the effectiveness of lipschitz-driven rehearsal in continual learning},
  author={Bonicelli, Lorenzo and Boschini, Matteo and Porrello, Angelo and Spampinato, Concetto and Calderara, Simone},
  journal={Advances in Neural Information Processing Systems},
  volume={35},
  pages={31886--31901},
  year={2022}
}

@misc{vicuna2023,
    title = {Vicuna: An Open-Source Chatbot Impressing GPT-4 with 90\%* ChatGPT Quality},
    url = {https://lmsys.org/blog/2023-03-30-vicuna/},
    author = {Chiang, Wei-Lin and Li, Zhuohan and Lin, Zi and Sheng, Ying and Wu, Zhanghao and Zhang, Hao and Zheng, Lianmin and Zhuang, Siyuan and Zhuang, Yonghao and Gonzalez, Joseph E. and Stoica, Ion and Xing, Eric P.},
    month = {March},
    year = {2023}
}

@article{lee2017overcoming,
  title={Overcoming catastrophic forgetting by incremental moment matching},
  author={Lee, Sang-Woo and Kim, Jin-Hwa and Jun, Jaehyun and Ha, Jung-Woo and Zhang, Byoung-Tak},
  journal={Advances in neural information processing systems},
  volume={30},
  year={2017}
}

@incollection{mccloskey1989catastrophic,
  title={Catastrophic interference in connectionist networks: The sequential learning problem},
  author={McCloskey, Michael and Cohen, Neal J},
  booktitle={Psychology of learning and motivation},
  volume={24},
  pages={109--165},
  year={1989},
  publisher={Elsevier}
}

@inproceedings{feng2022overcoming,
  title={Overcoming catastrophic forgetting in incremental object detection via elastic response distillation},
  author={Feng, Tao and Wang, Mang and Yuan, Hangjie},
  booktitle={Proceedings of the IEEE/CVF Conference on Computer Vision and Pattern Recognition},
  pages={9427--9436},
  year={2022}
}

@inproceedings{chen2023dynamic,
  title={Dynamic residual classifier for class incremental learning},
  author={Chen, Xiuwei and Chang, Xiaobin},
  booktitle={Proceedings of the IEEE/CVF International Conference on Computer Vision},
  pages={18743--18752},
  year={2023}
}

@inproceedings{lin2023pcr,
  title={PCR: Proxy-based contrastive replay for online class-incremental continual learning},
  author={Lin, Huiwei and Zhang, Baoquan and Feng, Shanshan and Li, Xutao and Ye, Yunming},
  booktitle={Proceedings of the IEEE/CVF Conference on Computer Vision and Pattern Recognition},
  pages={24246--24255},
  year={2023}
}

@article{tong2022videomae,
  title={Videomae: Masked autoencoders are data-efficient learners for self-supervised video pre-training},
  author={Tong, Zhan and Song, Yibing and Wang, Jue and Wang, Limin},
  journal={Advances in neural information processing systems},
  volume={35},
  pages={10078--10093},
  year={2022}
}

@inproceedings{khachatryan2023text2video,
  title={Text2video-zero: Text-to-image diffusion models are zero-shot video generators},
  author={Khachatryan, Levon and Movsisyan, Andranik and Tadevosyan, Vahram and Henschel, Roberto and Wang, Zhangyang and Navasardyan, Shant and Shi, Humphrey},
  booktitle={Proceedings of the IEEE/CVF International Conference on Computer Vision},
  pages={15954--15964},
  year={2023}
}

@inproceedings{dosovitskiy2020image,
  title={An Image is Worth 16x16 Words: Transformers for Image Recognition at Scale},
  author={Dosovitskiy, Alexey and Beyer, Lucas and Kolesnikov, Alexander and Weissenborn, Dirk and Zhai, Xiaohua and Unterthiner, Thomas and Dehghani, Mostafa and Minderer, Matthias and Heigold, Georg and Gelly, Sylvain and others},
  booktitle={International Conference on Learning Representations},
  year={2020}
}

@inproceedings{yang2024CLL-CLIP,
  title = {Embracing Language Inclusivity and Diversity in CLIP through Continual Language Learning},
  booktitle = {Proceedings of the AAAI Conference on Artificial Intelligence},
  author = {Yang, Bang and Dai, Yong and Cheng, Xuxin and Li, Yaowei and Raza, Asif and Zou, Yuexian},
  year = {2024},
  volume = {38},
  pages = {6458--6466},
}

@inproceedings{li2024kc,
  title={KC-Prompt: End-To-End Knowledge-Complementary Prompting for Rehearsal-Free Continual Learning},
  author={Li, Yaowei and Liu, Yating and Cheng, Xuxin and Zhu, Zhihong and Li, HongXiang and Yang, Bang and Huang, Zhiqi},
  booktitle={Proceedings of the IEEE International Conference on Acoustics, Speech and Signal Processing},
  pages={1--5},
  year={2024},
  organization={IEEE}
}

@inproceedings{yoon2018DEN,
  title = {Lifelong Learning with Dynamically Expandable Networks},
  booktitle = {International Conference on Learning Representations},
  author = {Yoon, Jaehong and Yang, Eunho and Lee, Jeongtae and Hwang, Sung Ju},
  year = {2018},
}

@inproceedings{li2019Learn2Grow,
  title = {Learn to Grow: A Continual Structure Learning Framework for Overcoming Catastrophic Forgetting},
  booktitle = {International Conference on Machine Learning},
  author = {Li, Xilai and Zhou, Yingbo and Wu, Tianfu and Socher, Richard and Xiong, Caiming},
  year = {2019},
  pages = {3925--3934}
}

@inproceedings{ke2020CAT,
  title = {Continual Learning of a Mixed Sequence of Similar and Dissimilar Tasks},
  booktitle = {Advances in neural information processing systems},
  author = {Ke, Zixuan and Liu, Bing and Huang, Xingchang},
  year = {2020},
  pages = {18493--18504}
}

@article{liu2023llava,
  title={Llava-plus: Learning to use tools for creating multimodal agents},
  author={Liu, Shilong and Cheng, Hao and Liu, Haotian and Zhang, Hao and Li, Feng and Ren, Tianhe and Zou, Xueyan and Yang, Jianwei and Su, Hang and Zhu, Jun and others},
  journal={arXiv preprint arXiv:2311.05437},
  year={2023}
}
\bibliographystyle{tmlr}

\appendix
\section{Appendix}
The outline of the appendix is as follows:
\begin{itemize}[topsep=0pt, partopsep=0pt, leftmargin=15pt, parsep=0pt, itemsep=3pt]
   \item Dataset Details
   \begin{itemize}
     \item Dataset Structure of VideoTool
     \item Prompt for Curating VideoTool
     \item Instruction Tuning Dataset Reformat
     \item Quality Control of VideoTool
   \end{itemize}
   \item Illustrations of Tool Selection
   \item Experimental Results
   \item Qualitative Results
 \end{itemize}

\subsection{Dataset Details} \label{sec:appendix1}

\textbf{Dataset Structure of VideoTool.} Each instance in VideoTool contains two rounds of conversation between \texttt{human} and \texttt{gpt}. The response of \texttt{gpt} includes three fields of \texttt{thought}, \texttt{action} and \texttt{value}. One example is available in Table \ref{table:incontextExample}.
\begin{table*}[ht]
        \caption{\textbf{Dataset structure of VideoTool.}}
        \renewcommand\arraystretch{1.1}
	\centering  
	\vspace{-2mm}
	\scalebox{0.9}{
			\begin{tabular}{lp{12cm}}
				\toprule
				\multicolumn{2}{l}{\bf An example of VideoTool dataset}  \\
				\midrule
				\multicolumn{2}{l}{\emph{\textbf{\textcolor{mygray}{First Round of Conversation}}}} \\
				Human & Please segment the objects in the video.\\
				GPT &$<$thoughts$>$ \\
				& Request a process of video object segmentation based on the user's prompt. \\
				& $<$actions$>$ \\
				& [\{`API$\_$name': `video-object-segmentation', `API$\_$params': \{\}\}]\\
				&$<$values$>$ \\
				& Sure thing! I'll initiate a object segmentation model to fulfill your request. Please wait while the processing takes place. \\
				\multicolumn{2}{l}{\emph{\textbf{\textcolor{mygray}{Second Round of Conversation}}}} \\
				Human & You have successfully completed the process of video captioning and object segmentation.\\
				GPT & $<$thoughts$>$ \\
				& I have successfully completed the process of video object segmentation. Now, it is time to update the user. \\
				& $<$actions$>$ \\
				& [] \\
				& $<$value$>$ \\
				& The objects have been segmented. You can now review the results. \\
				\bottomrule
			\end{tabular}
		}
		\label{table:incontextExample}   
	\end{table*}

\textbf{Prompt for Curating VideoTool.} We collect VideoTool by self-instruct GPT-3.5-turbo. The prompt and in-context learning cases are shown in Table \ref{table:prompt}.
\begin{table*}
\small
	\caption{\textbf{System prompts and in-context learning examples} for the generation of VideoTool.}
      \renewcommand\arraystretch{1.1}
		\centering  
		\vspace{-2mm}
                 \begin{tabular}{p{15cm}}
				\toprule
				\textbf{Prompts for GPT-3.5-turbo to generate instruction-tuning dataset:}  \\
				\midrule You are an AI assistant and you receive a set of conversations in json string format, the content of which is used as data for instruction finetune. Your task is to rephrase the dialogue to generate a new example. Note to give it in standard json format. DO NOT modify the ``from" and ``actions" parts. Only modify the ``value" and ``thoughts" part. \\
                    \midrule
    		    \textbf{In-context learning examples:}  \\
                    \midrule
                    \{"from": "human", "value": "Transcribe the audio content from the provided video.", "actions": []\}, \{"from": "gpt", "thoughts": "I need to find an ASR model and initiate the transcription process as per the user's request.", "actions": [\{"API$\_$name": "asr", "API$\_$params": \{\}\}], "value": "Certainly! I'll employ an Automatic Speech Recognition (ASR) model. Please wait while the ASR model processes the video"\}, \{"from": "human", "value": "asr output: You have used an ASR model to transcribe the audio content from the provided video based on my original request.", "actions": []\}, \{"from": "gpt", "thoughts": "Now that the ASR model has processed the video, I can update the user on the transcription results.", "actions": [], "value": "The ASR transcription you requested has been completed. Here is the transcribed text for your reference."\} \\

				\bottomrule
			\end{tabular}
		\label{table:prompt}  
\end{table*}



\textbf{Instruction Tuning Dataset Reformat.} In the second and third stage of COLT training, we reformat the instruction tuning dataset from LLaVA v1.5 \citep{liu2023improved} and Video-ChatGPT \citep{maaz2023video} into the \texttt{thought}-\texttt{action}-\texttt{value} pattern for the data format conformity. Specifically, the \texttt{thought} is generated by GPT indicating that the question can be answered without invoking any tools: 
\begin{adjustwidth}{2.5cm}{2cm}
\emph{The questions can be answered by the information in the context, without need any external tools.}
\end{adjustwidth}
The \texttt{action} is an empty list while \texttt{value} is the original response.

\textbf{Quality Control of VideoTool:} To ensure the reliability and integrity of the proposed dataset, we conduct quality control from two primary aspects: \emph{data format} checks and of \emph{semantic meaning} checks. The format check entails verifying whether the generated data adheres to a predefined structure, specifically confirming that each entry corresponds to a two-round conversation and that the responses from the GPT model contain essential components such as \texttt{thought}/\texttt{action}/\texttt{value} keys. Furthermore, this phase involves confirming the consistency of tool names used throughout the dataset. Manual checks on semantic meanings are conducted to validate the contextual relevance and accuracy of the generated content, ensuring that it aligns with the intended purpose of the dataset and maintains coherence within the conversations. 


\subsection{Illustrations of Tool Selection} \label{sec:appendix2}
\begin{figure}[t]
    \centering
    \includegraphics[width=\linewidth]{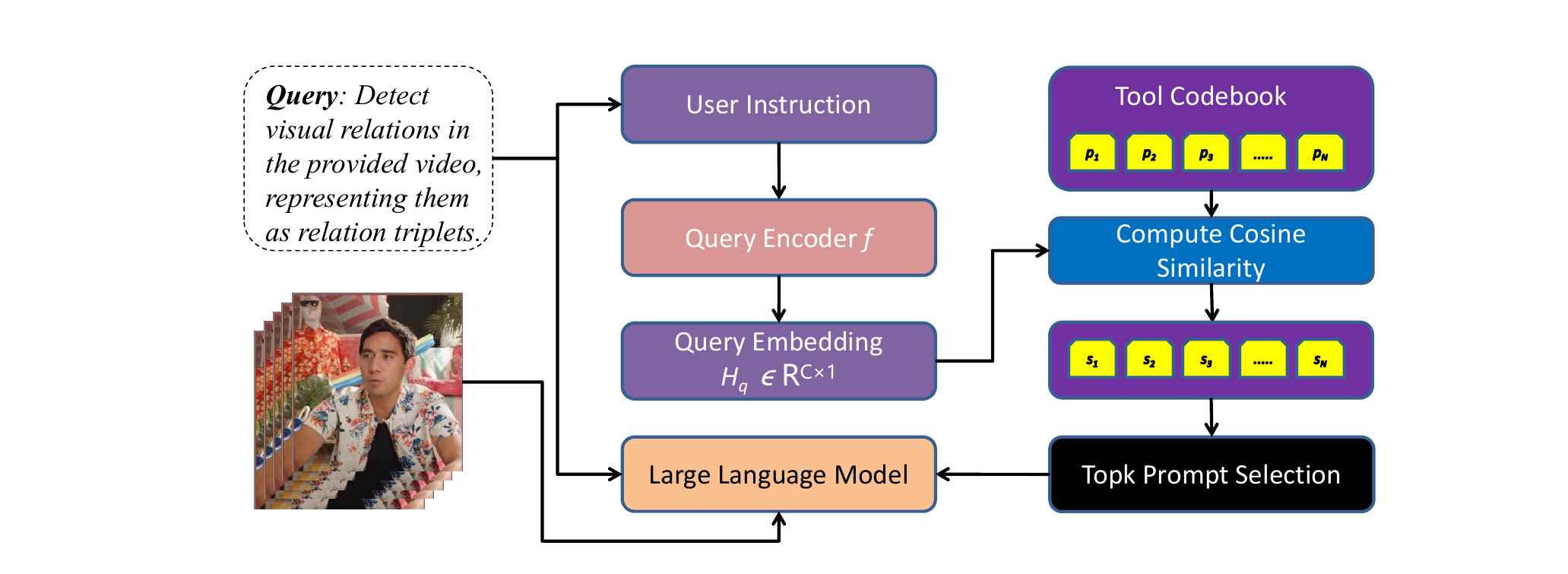}
    \caption{
    Illustration of the tool selection mechanism based on cosine similarity.
    Given the query embedding $\mathbf{h}_q$ and a tool codebook 
    $\{\mathbf{p}_1, \dots, \mathbf{p}_N\}$, we first compute the cosine 
    similarities between $\mathbf{h}_q$ and each tool prompt 
    $\mathbf{p}_i$. The most relevant tools are then selected (e.g., via 
    top-$K$) and concatenated with the visual/text embeddings before being
    fed into the large language model.
    }
    \label{fig:tool-selection}
\end{figure}

Figure~\ref{fig:tool-selection} shows the full pipeline of our tool selection mechanism.

\subsection{Experimental Results} \label{sec:appendix3}

In Sec.~\ref{sec:4_2}, we employ an additional CLIP-initialized  encoder ${f}_{\boldsymbol{\beta}}$ to extract query feature $\mathbf{H}_{\mathbf{q}}$. Then in Eq.~\eqref{eq:promptSelect}, the appropriate tool prompt is selected based on the cosine similarity with $\mathbf{H}_{\mathbf{q}}$. Here we conduct the ablation study which directly uses the encoded LLM features $\mathbf{H}_{\mathbf{w}}$ for similarity computation. The comparison results in Table \ref{tab:ablateQuery} demonstrate that the introduced CLIP feature $\mathbf{H}_{\mathbf{q}}$ leads to better performance. In addition, we further analyze how tool-use training and the codebook mechanism affect general video understanding beyond tool-centric benchmarks. 
As reported in Table~\ref{tab:ablate_vqa_codebook}, removing VideoToolBench data results in performance close to standard video-VQA models, while discarding the codebook causes substantial degradation across all benchmarks. 
These results suggest that the codebook plays a critical role in preserving and reusing tool-related knowledge, enabling COLT to transfer tool-use behaviors to standard VQA tasks rather than merely memorizing tool traces.
\begin{table}[t] \centering \small \setlength{\tabcolsep}{5pt} \begin{tabular}{lcccccc} \toprule \multirow{2}{*}{Method} & \multicolumn{2}{c}{MSVD-QA} & \multicolumn{2}{c}{MSRVTT-QA} & \multicolumn{2}{c}{ActivityNet-QA} \\ \cmidrule(lr){2-3}\cmidrule(lr){4-5}\cmidrule(lr){6-7} & Acc. $\uparrow$ & Score $\uparrow$ & Acc. $\uparrow$ & Score $\uparrow$ & Acc. $\uparrow$ & Score $\uparrow$ \\ \midrule COLT (w/o VideoToolBench) & 70.9 & 3.8 & 58.9 & 3.4 & 45.4 & 3.2 \\ COLT (w/o Codebook)       & 52.3 & 3.3 & 42.7 & 3.0 & 33.7 & 2.9 \\ COLT-joint                & \textbf{78.2} & \textbf{4.2} & \textbf{65.1} & \textbf{3.6} & \textbf{54.7} & \textbf{3.8} \\ \bottomrule \end{tabular} \caption{Controlled ablations on standard VQA benchmarks to disentangle the effects of tool-use training (VideoToolBench) and the codebook mechanism.} \label{tab:ablate_vqa_codebook} \end{table}
\begin{table*}[h]
	\small
	\setlength{\abovecaptionskip}{-0.05cm}
	\caption{\textbf{Ablations of the query feature source for tool prompt selection.}}
	\label{tab:ablateQuery}
	\renewcommand\arraystretch{1.1}
	\begin{center}
                \begin{tabular}{x{120}x{35}}
				\toprule
				\textbf{Query Feature Source} & \textbf{AVG} \\
				\midrule
				LLM feature $\mathbf{H}_{\mathbf{w}}$ & 47.4 \\
                    CLIP feature $\mathbf{H}_{\mathbf{q}}$ & \textbf{51.8} \\
				\bottomrule
			\end{tabular}
	\end{center}
\end{table*}

\subsection{Qualitative Results}

\begin{figure*}[t]
        \centering
         \includegraphics[width=0.65\textwidth]{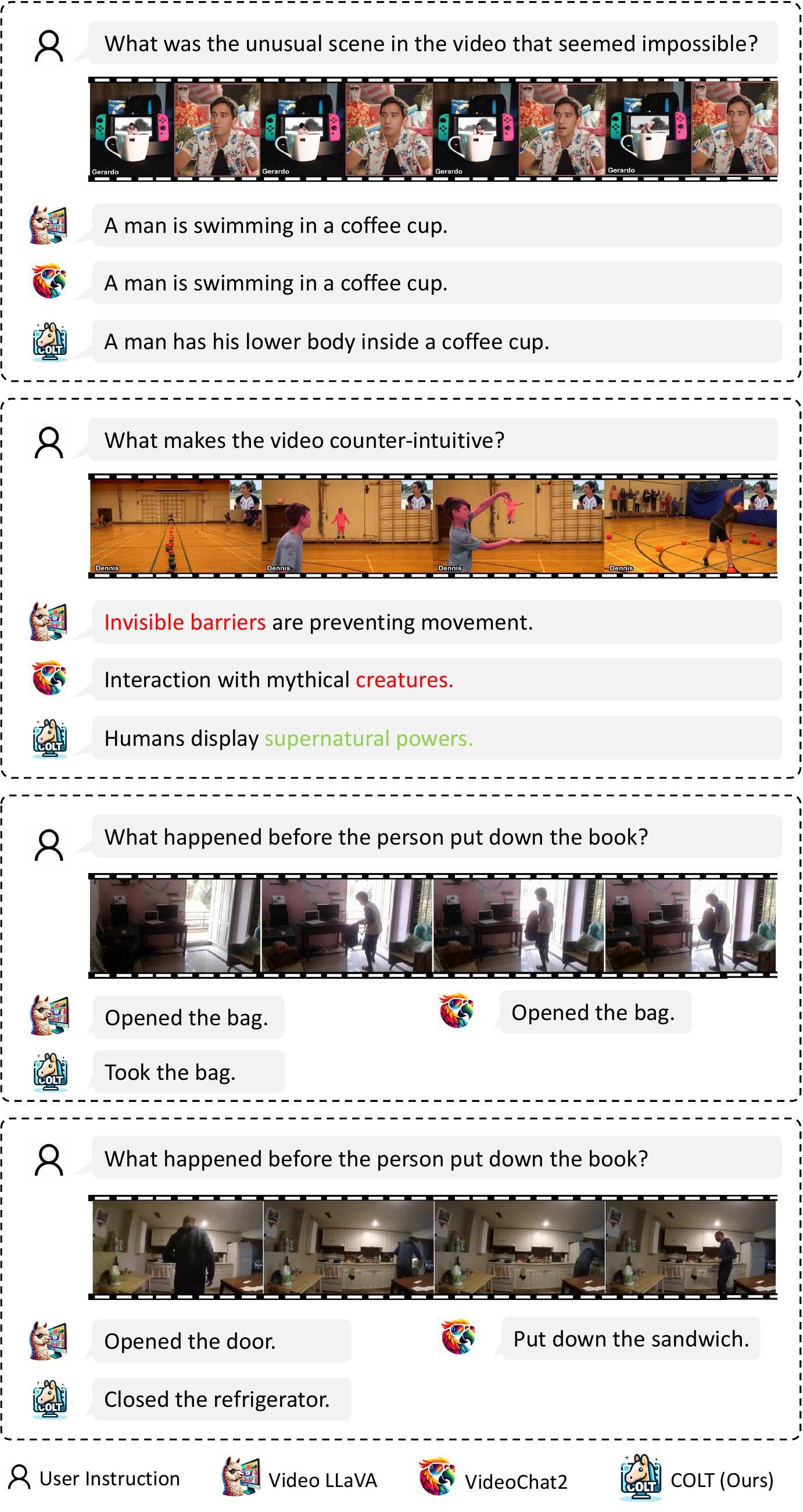} 
        \caption{\textbf{Qualitative results} on MVBench.}
	\label{fig:vis_MVBench}
\end{figure*}

\begin{figure*}[t]
        \centering
         \includegraphics[width=0.76\textwidth]{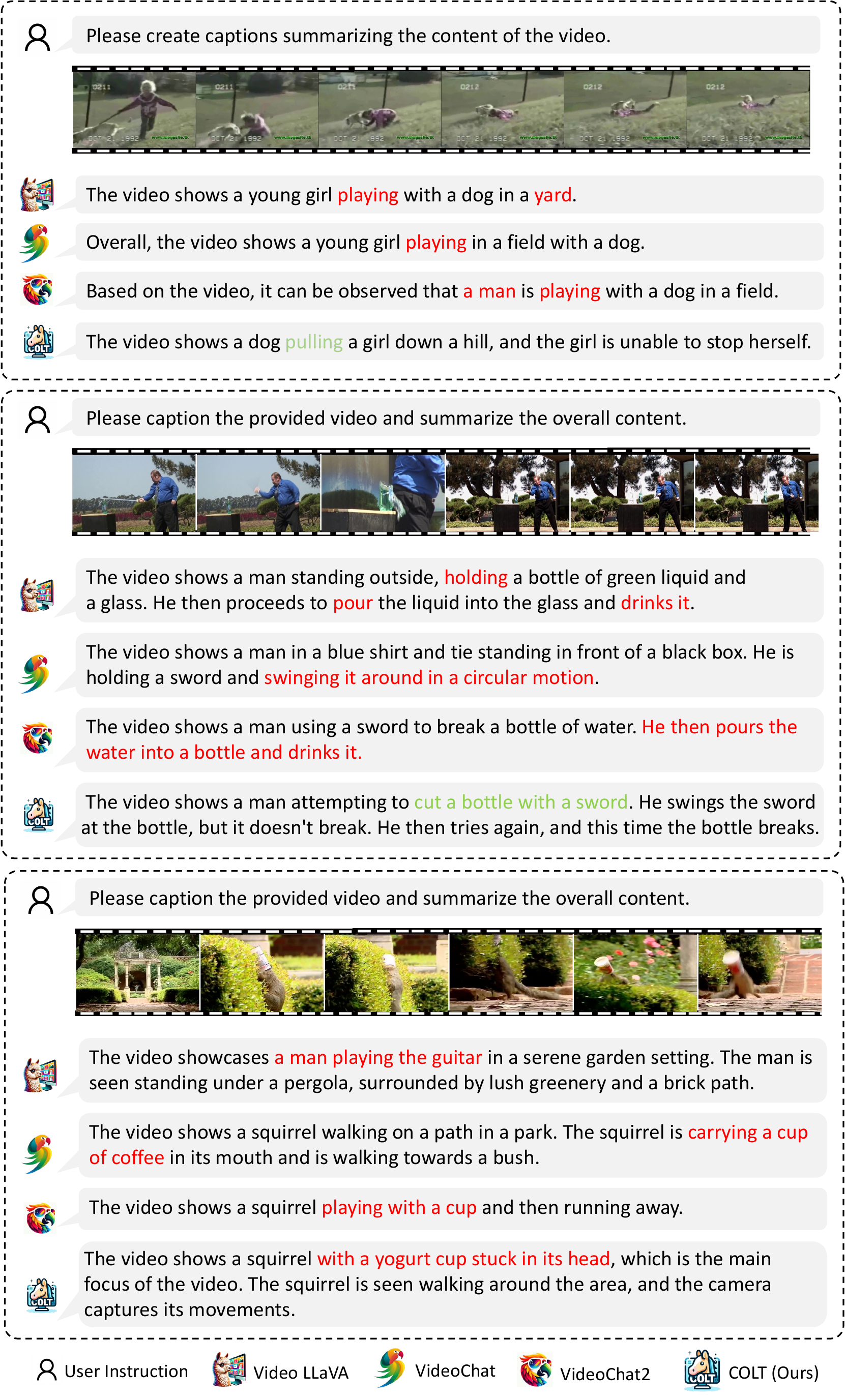} 
        \caption{\textbf{Qualitative results} on zero-shot video-question answering.}
	\label{fig:vis_qa}
\end{figure*}

\begin{figure*}[t]
        \centering
         \includegraphics[width=0.75\textwidth]{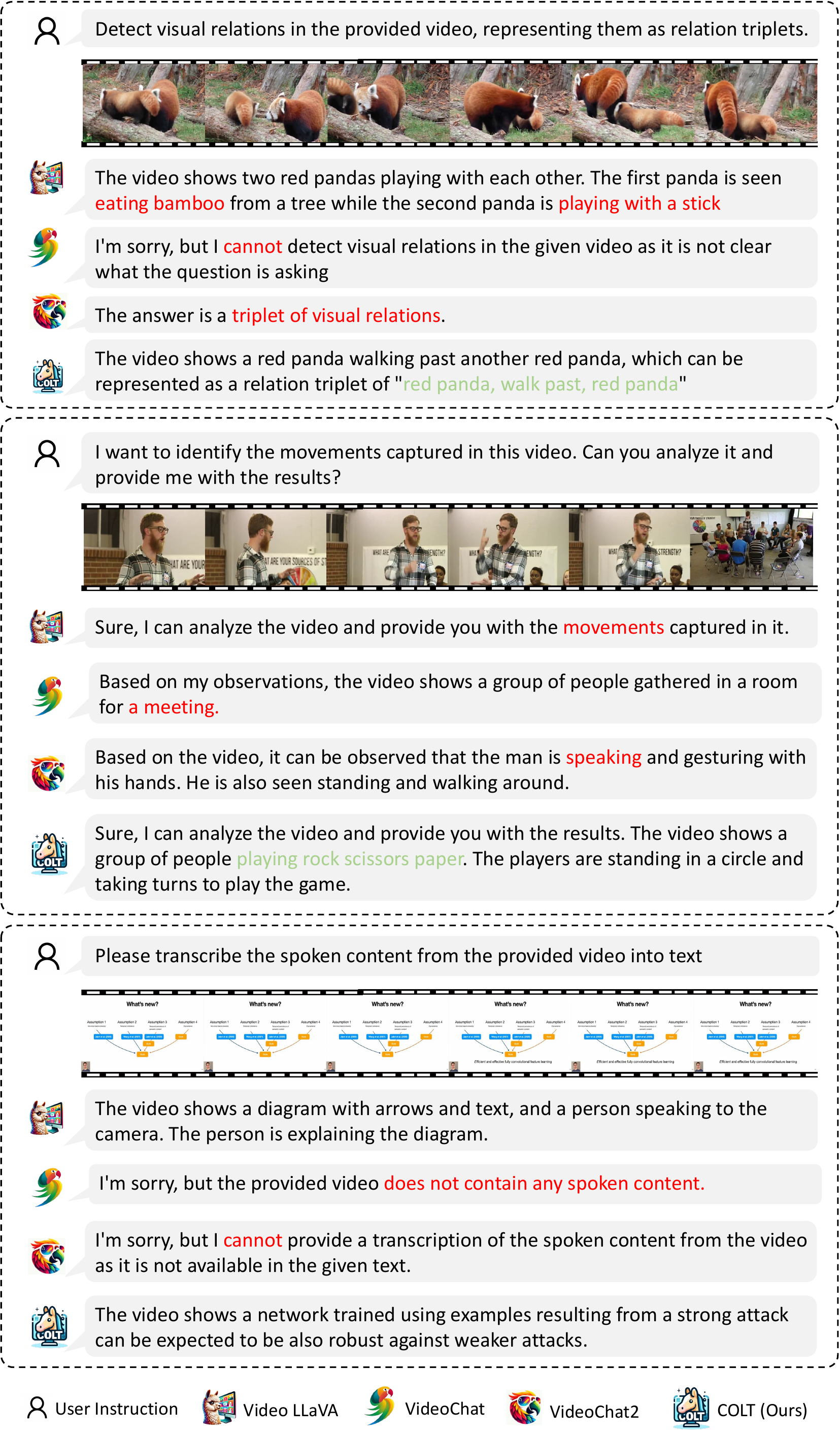} 
        \caption{\textbf{Qualitative results} on the test split of VideoTool.}
	\label{fig:vis_tool}
\end{figure*}

We present additional qualitative results on MVBench \citep{li2023mvbench}, zero-shot video-question answering \citep{chen2011collecting,xu2016msr,yu2019activitynet}, and VideoTool test split in Figure \ref{fig:vis_MVBench}, Figure \ref{fig:vis_qa}, and Figure \ref{fig:vis_tool}, respectively. Notably, our COLT precisely captures dynamic video information and generate more reasonable responses.

\end{document}